\documentclass{ecai} 

\usepackage{latexsym}
\usepackage{amssymb}
\usepackage{amsmath}
\usepackage{amsthm}
\usepackage{booktabs}
\usepackage{nicematrix}
\usepackage{hhline}
\usepackage{amsmath,amssymb} 
\usepackage{algorithm}
\usepackage{algorithmicx}
\usepackage{algpseudocode}
\usepackage{graphics}
\usepackage{threeparttable}
\usepackage{color}
\usepackage[normalem]{ulem}
\usepackage{multirow}
\usepackage{float}
\usepackage{amsfonts}
\usepackage{bm}
\usepackage{array}
\usepackage{pifont}
\usepackage{diagbox}
\usepackage{rotating}
\usepackage{booktabs}
\usepackage{overpic}
\usepackage{textcomp}
\usepackage{contour}

\usepackage{csquotes}
\usepackage[american]{babel}
\usepackage{microtype}
\usepackage{bbding}
\usepackage{amsfonts,amssymb}
\usepackage[pagebackref=false,breaklinks=true,colorlinks, bookmarks=false]{hyperref}
\usepackage{colortbl}
\usepackage{cleveref}
\usepackage{appendix}
\usepackage{silence}
\usepackage{booktabs}
\usepackage{makecell}
\def\ie{\emph{i.e.}}
\def\eg{\emph{e.g.}}
\def\etc{\emph{etc}}

\definecolor{xj}{RGB}{0,0,255}

\definecolor{light-gray}{gray}{0.5}

\def\ie{\emph{i.e.}}
\def\eg{\emph{e.g.}}
\def\etc{\emph{etc}}

\crefformat{section}{\S#2#1#3} 
\crefformat{subsection}{\S#2#1#3}
\crefformat{subsubsection}{\S#2#1#3}

\newcommand{\BibTeX}{B\kern-.05em{\sc i\kern-.025em b}\kern-.08em\TeX}

\begin{document}
\begin{frontmatter}

\title{Task-Aware Dynamic Transformer for 
Efficient Arbitrary-Scale Image Super-Resolution}

\author[A]{\fnms{Tianyi}~\snm{Xu}}
\author[A]{\fnms{Yijie}~\snm{Zhou}}
\author[B]{\fnms{Xiaotao}~\snm{Hu}}
\author[C]{\fnms{Kai}~\snm{Zhang}}
\author[D]{\fnms{Anran}~\snm{Zhang}}
\author[E,F]{\fnms{Xingye}~\snm{Qiu}}
\author[A,G]{\fnms{Jun}~\snm{Xu}\thanks{Corresponding Author. Email: csjunxu@nankai.edu.cn}} 

\address[A]{\fontsize{10pt}{12pt} \selectfont School of Statistics and Data Science, Nankai University, Tianjin, China } 
\address[B]{\fontsize{10pt}{12pt} \selectfont College of Computer Science, Nankai University, Tianjin, China } 
\address[C]{\fontsize{10pt}{12pt} \selectfont School of Intelligence Science and Technology, Nanjing University, Suzhou, China}    
\address[D]{\fontsize{10pt}{12pt} \selectfont Tencent Data Platform, Beijing, China} 
\address[E]{\fontsize{10pt}{12pt} \selectfont Zhejiang University, Hangzhou, China}  
\address[F]{\fontsize{10pt}{12pt} \selectfont Systems Engineering Research Institute, China State Shipbuilding Corporation Limited, Beijing, China}
\address[G]{\fontsize{10pt}{12pt} \selectfont Guangdong Provincial Key Laboratory of Big Data Computing, The Chinese University of Hong Kong, Shenzhen, China} 

\begin{abstract}
Arbitrary-scale super-resolution (ASSR) aims to learn a single  model for image super-resolution at arbitrary magnifying scales.
Existing ASSR networks typically comprise an off-the-shelf scale-agnostic feature extractor and an arbitrary scale upsampler. 
These feature extractors often use fixed network architectures to address different ASSR inference \textbf{tasks}, each of which is characterized by an \textbf{input} image and an upsampling \textbf{scale}.
However, this overlooks the difficulty variance of super-resolution on different inference scenarios, where simple images or small SR scales could be resolved with less computational effort than difficult images or large SR scales.
To tackle this difficulty variability, in this paper, we propose a Task-Aware Dynamic Transformer (TADT) as an input-adaptive feature extractor for efficient image ASSR.
Our TADT consists of  a multi-scale feature extraction backbone built upon groups of Multi-Scale Transformer Blocks (MSTBs) and a Task-Aware Routing Controller (TARC).
%
The TARC predicts the inference paths within feature extraction backbone, 
specifically selecting MSTBs based on the input images and SR scales.
The prediction of inference path is guided by a new loss function to trade-off the SR accuracy and efficiency.
Experiments demonstrate that, when working with three popular arbitrary-scale upsamplers, our TADT achieves state-of-the-art ASSR performance when compared with mainstream feature extractors, but with relatively fewer computational costs.
%
The code is available at \url{https://github.com/Tillyhere/TADT}.
\end{abstract}

\end{frontmatter}


\section{Introduction}

The goal of Arbitrary-Scale Super-Resolution (ASSR) is to learn a single model capable of performing image super-resolution at arbitrary scales~\cite{hu2019meta,2020liif,wang2021arbsr,yang2021implicit}.
Current ASSR methods~\cite{hu2019meta,2020liif,ciao_sr,lte-jaewon-lee} primarily focus on developing arbitrary-scale upsamplers to predict the high-resolution (HR) image from the feature of a low-resolution (LR) image extracted by an off-the-shelf feature extractor~\cite{2017edsr,zhang2018rdn,liang2021swinir}.
Inspired by the merits of meta-learning~\cite{fin2017icml}, the work of MetaSR~\cite{hu2019meta} learns adaptive upsamplers according to the SR scale.
Later, the methods of~\cite{2020liif,yang2021implicit,lte-jaewon-lee,Chen_2023_CVPR} leverage the implicit neural representation~\cite{2018implicit} to predict the upsampled HR image by the coordinates and the feature map of the corresponding LR image.
However,  feature extractors~\cite{2017edsr,zhang2018rdn,liang2021swinir} in these ASSR methods are usually scale-agnostic, which discourage the adaptivity of extracted feature map to the multiple user-defined SR scale and leads to inferior SR results~\cite{wang2021arbsr,idm_Gao_2023_CVPR}.


\begin{figure}[!tbp]\footnotesize 
 \hspace{-0.8cm}
\begin{tabular}{c@{\extracolsep{0.0em}}@{\extracolsep{0.01em}}c}
      {\begin{overpic}[width=0.53\linewidth,height=0.38\linewidth]{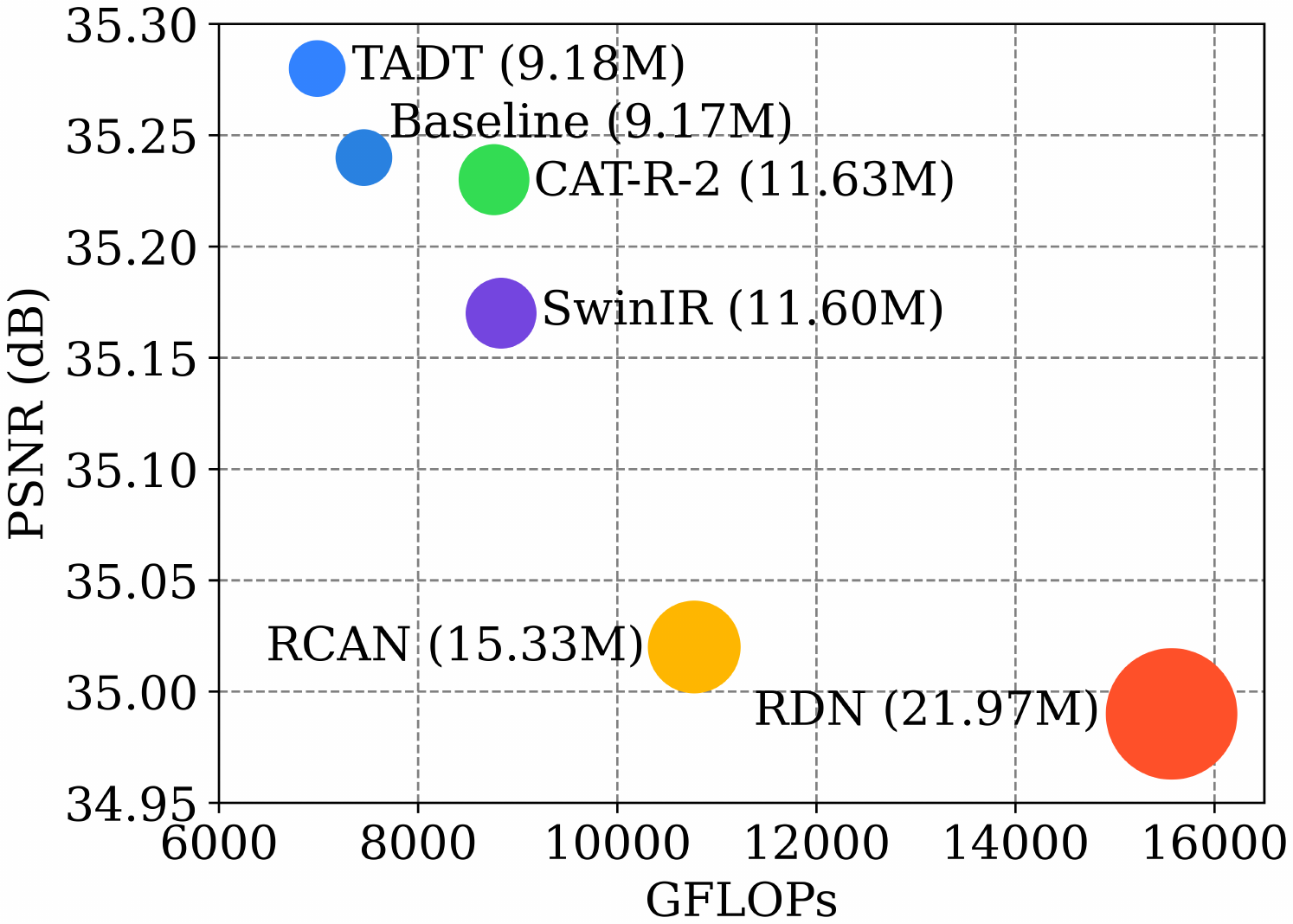}
\end{overpic}} & {\begin{overpic}[width=0.53\linewidth,height=0.38\linewidth]{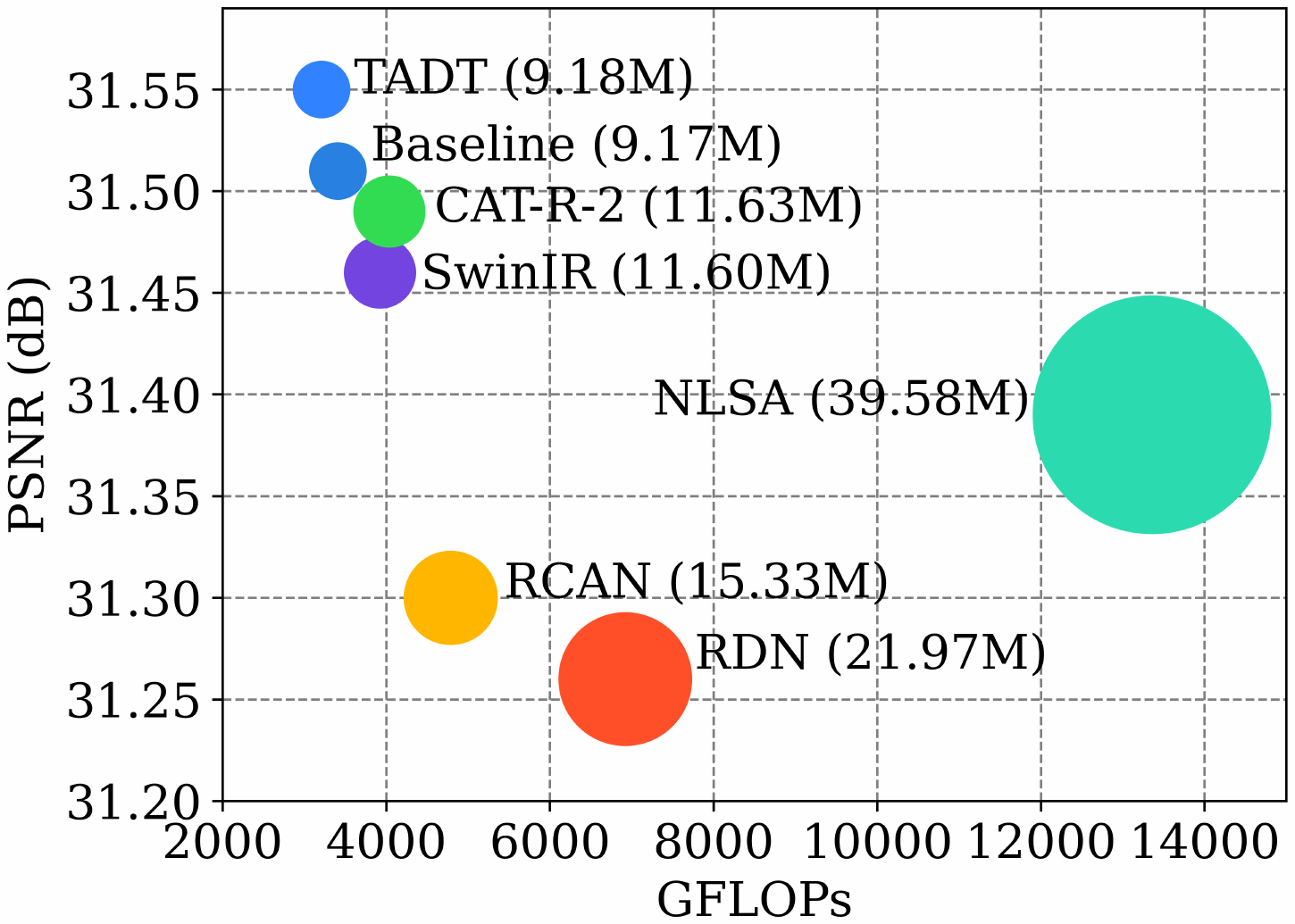}
\end{overpic}} 
  \\
 {\ \ \ \ \ \ (a)} & {\ \ \ \ \ \ (b)}  \\

\end{tabular}
\caption{\textbf{PSNR results, 
GFLOPs
and parameter amounts of different feature extractors} working with the upsampler LIIF~\cite{2020liif} for (a) $\times 2$ super-resolution and (b) $\times 3$ super-resolution on the DIV2K validation set~\cite{Agustsson_2017_CVPR_Workshops}.
The disc radius is proportional to parameter amounts of different feature extractors.
%
Baseline and TADT
are our feature extractors that will be introduced in \S\ref{sec:tadt}.
}
\label{fig:intro}
 \vspace{3mm}
\end{figure}

To extract scale-aware feature for image ASSR, some researchers attempt to design scale-conditional convolutions to dynamically generate scale-aware filters~\cite{wang2021arbsr,idm_Gao_2023_CVPR,EQSR}.
For example, ArbSR~\cite{wang2021arbsr} employs scale-aware convolution which fuses a set of filters using  weights dynamically generated based on the scale information.
The method of EQSR~\cite{EQSR} achieves adaptive modulation of convolution kernels with scale-aware embedding.
Implicit Diffusion Model~\cite{idm_Gao_2023_CVPR} presents a scale-aware mechanism to work with a denoising diffusion model for high-fidelity image ASSR.
In short, these methods mainly implement feature-level or parameter-level modulation mechanisms for scale-aware feature extraction.
However, these feature extractors tackle the input images and SR scales of different difficulty by fixed network architectures.
This inevitably emerges huge computational redundancy in ``easy'' inference scenarios, \eg, ``simple'' images and/or small SR scales, that can be effectively resolved with less computational effort.

The variability of restoring difficulty on different images is inherent in image restoration~\cite{classsr}.
It is more evident for image ASSR, 
since  the difficulty variance of ASSR
comes from not only  content-diverse input images, but also different upsampling scales.
On one hand, content-diverse images often suffer from different restoration difficulty and require image-adaptive inference complexity~\cite{classsr,hu2022mga}.
On the other, the difficulty variability of image ASSR emerges as higher upsampling scales usually need larger computational burden 
~\cite{2020deepSR_survey}.
%
%
Considering an input image and the corresponding upsampling scale factor as an ASSR task, it is essential to develop task-aware feature extractors with adaptive inference based on  the difficulty of ASSR tasks.

For this goal, in this work, we propose the Task-Aware Dynamic Transformer (TADT) as an efficient feature extractor, with dynamic computational graphs upon different ASSR tasks.
Specifically, our feature extractor TADT has a feature extraction backbone and a Task-Aware Routing Controller (TARC).
The backbone contains multiple Multi-Scale Transformer Blocks (MSTBs) to exploit multi-scale representation~\cite{Li_2018_ECCV,Zamir2021Restormer,zhang2022elan}.
Our TARC predicts the inference path of the backbone for each ASSR task, realizing task-aware inference with dynamic architectures.
It is a two-branch module to transform the input image and SR scale into a probability vector and an intensity indicator 
respectively.
%
%
The probability vector is modulated by the intensity indicator to produce the sampling probability vector, which is used to predict the final routing vector by Bernoulli Sampling combined with the Straight-Through Estimator~\cite{bengio2013estimating,hubara2017quantized}.
The routing vector determines the computational graph of the feature extraction backbone in our TADT to make it aware of input images and scales for image ASSR.

%
To make TADT more efficient,  we further design a loss function to penalize the intensity indicator.
Experiments on ASSR 
demonstrate that, TADT outperforms mainstream feature extractors by fewer parameter amounts and computation costs, and better ASSR performance when working with arbitrary-scale upsamplers of MetaSR~\cite{hu2019meta}, LIIF~\cite{2020liif}, and LTE~\cite{lte-jaewon-lee} (a glimpse is provided in Figure~\ref{fig:intro}).


Our main contributions can be summarized as follows:
\vspace{-3.5mm}
\begin{itemize}
\item We propose the Task-Aware Dynamic Transformer (TADT) as a new feature extractor for efficient image ASSR. The main backbone of our TADT is built upon cascaded multi-scale transformer blocks (MSTBs) to learn expressive feature representations.
\item 
We develop a task-aware routing controller to predict adaptive inference paths within the main backbone of  TADT feature extractor for different ASSR tasks defined by the input image and SR scale.
\item 
We devise an intensity loss function to guide the prediction of inference paths in our feature extraction backbone, leading to efficient image ASSR performance.
\end{itemize}

%
%

\section{Related Work}
\label{related work}
\subsection{Arbitrary-Scale Image Super-Resolution}
\label{sec:assr}
Arbitrary-Scale Super-Resolution (ASSR) methods learn a single SR model to tackle the image super-resolution of arbitrary scale factors~\cite{hu2019meta}.
Meta-SR~\cite{hu2019meta} 
represents one of the earliest endeavors in image ASSR,
which dynamically predicts  weights of filters for different scales by the meta-upscale module inspired by the meta-learning scheme~\cite{fin2017icml}.
Then, LIIF~\cite{2020liif} pioneers local implicit neural representation for continuous image upsampling.
Following this direction, Ultra-SR~\cite{ultrasr} integrates spatial encoding with implicit image function to improve the recovery of high-frequency textures. 
%
LTE~\cite{lte-jaewon-lee} 
transforms the spatial coordinates into the Fourier frequency domain and learns implicit representation for detail enhancement.
%
%
Attention~\cite{transformer2017} is also exploited in the methods of ITSRN, Ciao-SR~\cite{ciao_sr} and CLIT~\cite{Chen_2023_CVPR}.
These methods mainly focus on designing scale-aware upsamplers, but often employ input-agnostic feature extractors~\cite{2017edsr,zhang2018rdn,liang2021swinir} leading to inferior image ASSR performance~\cite{wang2021arbsr,ijcai2023p63,EQSR}.

To mitigate this, recent ASSR methods~\cite{wang2021arbsr,ijcai2023p63,EQSR} incorporate scale information into the feature extractors.
 ArbSR~\cite{wang2021arbsr} and EQSR~\cite{EQSR} dynamically predicts filter weights from scale-conditioned feature extraction.
Differently, LISR~\cite{ijcai2023p63} and IDM~\cite{idm_Gao_2023_CVPR} learn scale-conditioned attention weights to modulate scale-aware feature channels.
These methods mainly extract scale-aware feature by feature or parameter level modulation, but with fixed inference architectures.
This still limits their efficiency to tackle the versatile images and SR scale factors in ASSR.
In this work, we propose a hyper-network~\cite{ha2017hypernetworks} as the feature extractor that is aware of both the image and scale to achieve dynamic ASSR inference with adaptive computational efficiency.

\subsection{Dynamic Networks}
\label{sec:dnet}
The dynamic inference is explored mainly from three aspects for expressive representation power and adaptive inference efficiency~\cite{han2021dynamic}: spatially-adaptive~\cite{chen2016sca,Chen2020DynamicRC}, temporal-adaptive~\cite{NEURIPS2019_liteeval,ghodrati2021,actionspotter}, and sample-adaptive
~\cite{2017skipnet}.
By taking input image and SR scale as an inference sample, our Task-Aware Dynamic Transformer (TADT) based ASSR network belongs to the category of 
sample-adaptive dynamic inference.
Sample-adaptive dynamic networks have been developed mainly to learn dynamic parameters or architectures~\cite{han2021dynamic}.
Parameter-dynamic methods~\cite{deformable,ma2020weightnet} only tailor the network parameters according to the input, but under fixed network architectures.
Architecture-dynamic methods mainly perform inference from three aspects: dynamic depth
~\cite{2017msdnet},
dynamic width~\cite{dynamic_slimmable,10054501}, and dynamic routing~\cite{Liu2017DynamicDN,trar}.
The depth-dynamic inference mainly resort to early exiting~\cite{han2021dynamic} or layer skipping~\cite{2017skipnet}.
The width-dynamic inference~\cite{dynamic_slimmable,10054501} typically leverage dynamic channel or neuron pruning techniques~\cite{NIPS2017_pruning}.
Dynamic inference routing is usually employed to learn sample-specific inference architecture~\cite{Liu2017DynamicDN,trar}.
In this work, we develop a transformer-based multi-branch feature extractor, and arm it with a task-aware network routing controller for architecture-dynamic image ASSR inference.
\begin{figure}[t]
	\centering
 \begin{overpic}[width=\linewidth]
     {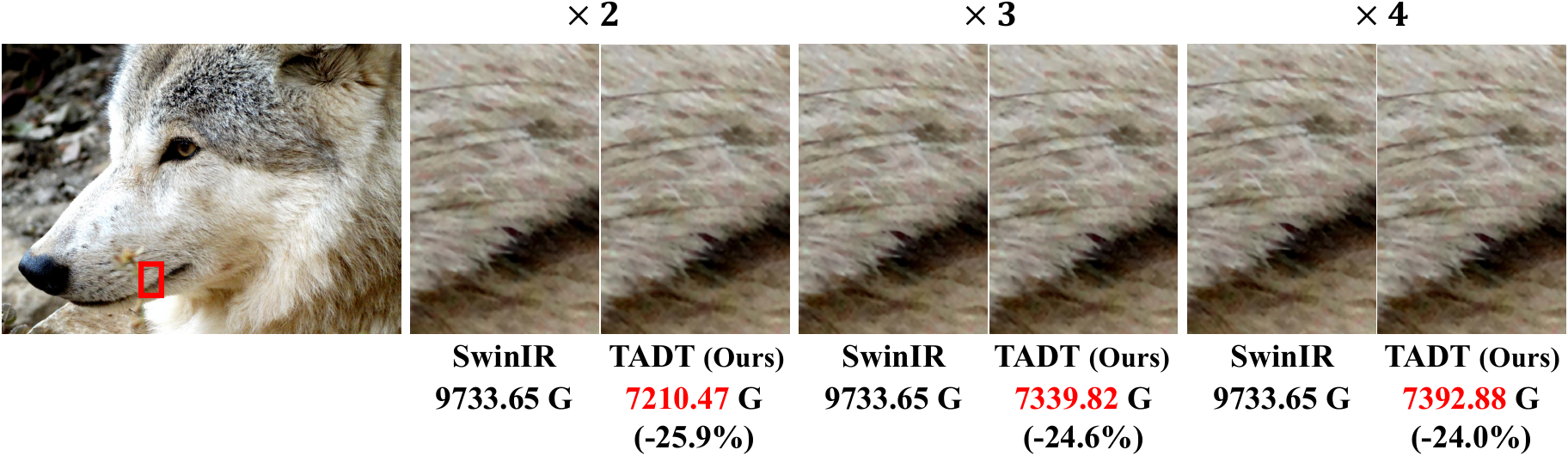}
     \put(2,3){\scriptsize $1020\times768$ Input}
 \end{overpic}
\vspace{-3.5mm}
   \caption{\textbf{Computational FLOPs of SwinIR~\cite{liang2021swinir} and our TADT} on one image from DIV2K at different SR scales. The arbitrary-scale upsampler is LIIF~\cite{2020liif}. Our TADT uses less computational costs for smaller SR scales.}
 \label{fig:motivation}
   \vspace{3mm}
\end{figure}  

\begin{figure*}[t]
	\centering
 \vspace{-5mm}
    \begin{overpic}[width=0.88\linewidth]{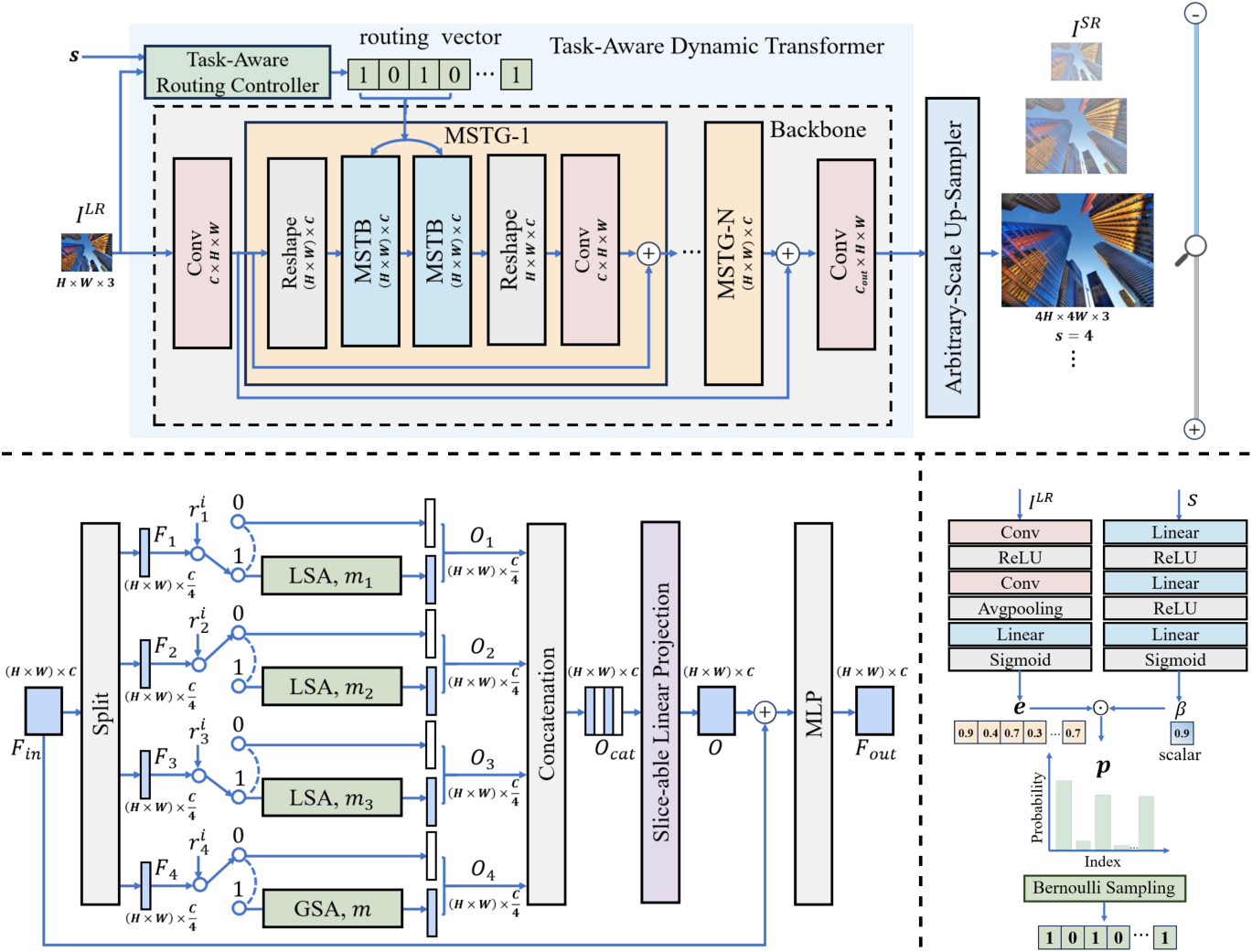}
\put(1.4,75.6){\footnotesize \textbf{(a) Architecture of Our ASSR Network}}
\put(1.4,38){\footnotesize \textbf{(b) Multi-Scale Transformer Block (MSTB)}}
\put(73.9,38){\footnotesize \textbf{(c) Task-Aware Routing Controller}}
\end{overpic}
\vspace{-1mm}
\caption{
\textbf{Illustration of our ASSR network}.
(a) Architecture of our ASSR network containing our Task-Aware Dynamic Transformer and an arbitrary-scale upsampler.
(b) Our Multi-scale Transformer Block (MSTB). ``LSA'' indicates local self-attention with a window size of $m_j$ for the $j$-th branch. ``GSA'' indicates global self-attention with a window size of $m$.
(c) Our Task-Aware Routing Controller is a two-branch module aware of the input image $\bm{I}^{LR}$ and SR scale $s$.
} 
 \vspace{1mm}
 \label{fig:overall}
\end{figure*}  

\section{Methodology}
\label{method}
%
%
\subsection{Motivation}
\label{sec:motivation}

Scale-agnostic feature extractors for ASSR consume the same computation overhead for super-resolution of different images or scales, and ignore the variance of super-resolution difficulty for diverse ASSR tasks (input images and SR scales)~\cite{2020deepSR_survey}.
This brings computational redundancy to these feature extractors upon relatively ``easy'' ASSR tasks.
To illustrate this point, in Figure~\ref{fig:motivation}, we compare the SR images of LIIF~\cite{2020liif} using SwinIR~\cite{liang2021swinir} or our method (will be introduced later) as the feature extractor on one $1020 \times 768$ image from the DIV2K dataset.
One can see that SwinIR needs a constant FLOPs of 9733.65G to extract features for ASSR task with SR scales of $\times 2$, $\times 3$, and $\times 4$.
On the contrary, our TADT needs less computational costs for SR tasks of lower scales, and enables LIIF~\cite{2020liif} to output SR images with similar or even better image quality than those of SwinIR.
In the end, it is promising to develop a feature extractor with dynamic computational graphs for image ASSR, which is the main motivation of our work.
Next, we will elaborate our method for image ASSR.

\subsection{Network Overview}
\label{sec:overview}
The overall pipeline of our ASSR network is illustrated in Figure~\ref{fig:overall} (a).
It takes our Task-Aware Dynamic Transformer (TADT) as the feature extractor and an arbitrary-scale upsampler to output the magnified image.
Our TADT extractor comprises a main multi-scale feature extraction backbone and a Task-Aware Routing Controller (TARC).
The feature extraction backbone first utilizes a convolution layer to obtain the shallow feature.
It then learns scale-aware deep feature, with the routing vector provided by our TARC, by $N$ cascaded Multi-Scale Transformer Groups (MSTGs) appended by a convolution layer.
Each MSTG group contains two Multi-Scale Transformer Blocks (MSTBs) and a convolution layer, and 
each MSTB learns multi-scale representation by four self-attention branches.
%
A skip connection is used to fuse the shallow feature and the extracted feature by $N$ MSTG groups.
Our TARC controller predicts the routing vector of our TADT feature extraction backbone, \ie, the selection of self-attention branches, for different input LR images and SR scales.
More detailed structure of our TADT will be presented in \S\ref{sec:tadt}.

For the arbitrary-scale upsampler, we employ the off-the-shelf methods such as MetaSR~\cite{hu2019meta}, LIIF~\cite{2020liif}, and LTE~\cite{lte-jaewon-lee}, \etc.
%

\subsection{Proposed Task-Aware Dynamic Transformer}
\label{sec:tadt} 
In this work, we propose a task-aware feature extractor based on transformers~\cite{liang2021swinir,zhang2022elan} for image ASSR.
The proposed extractor can adjust its computational graph according to different LR images and upsampling scales, to achieve dynamic feature extraction with adaptive computational costs.
Since each set of input LR image and upsampling scale constitute 
the inputs of an inference task in ASSR, our feature extractor is termed as Task-Aware Dynamic Transformer (TADT).

Given an inference task consisting of an LR image $\bm{I}^{LR}$ and an upsampling scale factor $s$, our Task-Aware Routing Controller (TARC) first predicts a binary routing vector $\bm{r}\in\{0,1\}^{4N}$.
Here, $4N$ is the number of controllable self-attention branches in the feature extraction backbone, since each MSTB block has four self-attention branches and the two MSTBs in each MSTG group use the same branches.
The backbone then encodes the LR image $\bm{I}^{LR}$ of an input task and determine its computational graph according to the routing vector $\bm{r}$. Specifically, the routing vector $\bm{r}$ consists of $N$ sets of 4-dimensional routing sub-vectors as $\bm{r} = \left[\bm{r}^1, \cdots, \bm{r}^i, \cdots \bm{r}^N\right]$, where $\bm{r}^i = \left[r_{1}^{i}, r_{2}^{i}, r_{3}^{i}, r_{4}^{i}\right]$.
Here, $\bm{r}^{i}$ ($i=1,...,N$) is the sub-vector of the $i$-th MSTG and 
$r_{j}^{i}=0 \text{ or } 1$ ($j\in\{1,2,3,4\}$)
is the routing index of the $j$-th self-attention branch.
$r_{j}^{i}=1$ means that the $j$-th branch of two MSTBs in the $i$-th MSTG is used. Otherwise, this branch will be bypassed.
Our experiments show that using separate routing sub-vectors for the two MSTBs in each MSTG achieves similar ASSR performance.
Thus, we share the same routing sub-vectors on the two MSTBs in each MSTG for model simplicity.

\begin{figure}[t]
\vspace{-5mm}
	\centering
     \begin{overpic}[width=0.9\linewidth]
     {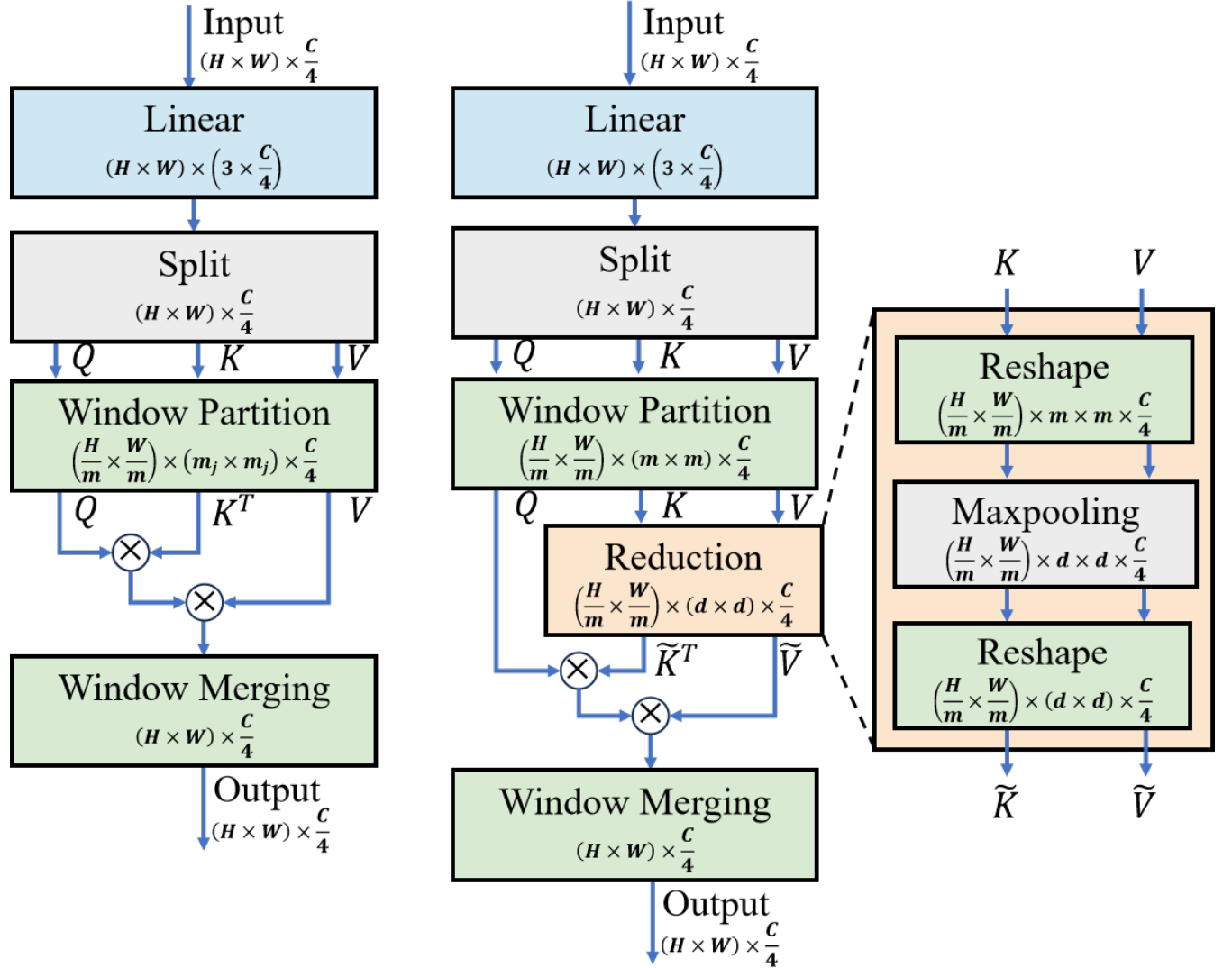}
    
\put(15.5,0){\small (a)}
\put(66,0){\small (b)}
     \end{overpic}
       \vspace{-1mm}
   \caption{Illustration of  (a) local self-attention  and  (b) global self-attention. }
 \label{fig:sa}
 \vspace{5mm}
\end{figure}  
\subsubsection{Multi-Scale Transformer Block}
\label{sec:mstb}

By leveraging the power of multi-scale learning~\cite{Li_2018_ECCV,Zamir2021Restormer,zhang2022elan} and global learning~\cite{xie2021segformer,tu2022maxvit}, we also propose a new Multi-Scale Transformer Block (MSTB) for comprehensive representation learning.
Take the MSTB in the $i$-th MSTG as an example.
As shown in Figure~\ref{fig:overall} (b), the MSTB block in each MSTG mainly has three local self-attention (LSA) branches with different window sizes \{$m_1$, $m_2$, $m_3$\} to learn abundant multi-scale representation and a global self-attention (GSA) to provide global insight.
It first splits the reshaped feature map $\bm{F}_{in}\in\mathbb{R}^{(H \times W) \times C}$ into four groups along the channel dimension, yielding $\left\{\bm{F}_j\right\}_{j=1}^4$ of size $(H \times W)\times \frac{C}{4}$.
The routing sub-vector $\bm{r}^i$ indicates the forward path of four split feature maps $\left\{\bm{F}_j\right\}_{j=1}^4$.
If the routing value $r_{j}^{i}=1$, the split feature map $\bm{F}_j$ will be fed into the $j$-th self-attention (LSA or GSA) branch.
Otherwise if $r_{j}^{i}=0$, the split feature map $\bm{F}_j$ will be set as a comfortable zero tensor and bypass the $j$-the attention branch.
The outcome split feature $\bm{O}_j$ of this process can be expressed as:
\vspace{-1mm}
\begin{equation}
\bm{O}_j=
\begin{cases}
\operatorname{SA}_j(\bm{F}_j), & r_{j}^{i} =1,
\\ 
\bm{0}, & r_{j}^{i} =0,
\end{cases}
\end{equation}
\vspace{0mm}
where $\operatorname{SA}_j$ is the $j$-th self-attention branch of this MSTB.

Subsequently, the outcome split features of four branches $\left\{\bm{O}_j\right\}_{j=1}^4$ are concatenated to obtain the outcome feature $\bm{O}_{cat}$.
The $\bm{O}_{cat}$ is further fed into our efficient slice-able linear projection.
The resulting outcome feature $\bm{O}$ is then added to the input feature $\bm{F}_{in}$, and further processed by a standard MLP in transformer blocks to output the feature $\bm{F}_{out}$ of this MSTB.

\noindent{\bf Local self-attention} (LSA).
As illustrated in Figure~\ref{fig:sa} (a), given an input feature of size $(H \times W) \times \frac{C}{4}$, the LSA branch first expands the channel dimension to $\frac{3C}{4}$ by a linear layer and then splits it along the channel dimension into a Query matrix $\bm{Q}$, a Key matrix $\bm{K}$, and a Value matrix $\bm{V}$, all of size $(H \times W) \times \frac{C}{4}$.
The local window attention partitions $\bm{Q}$, $\bm{K}$, $\bm{V}$ into windows of size $m_j \times m_j$ ($j=1,2,3$) and computes the attention map within each window.
After performing self-attention along the window dimension, the LSA branch reshapes 
the attention feature
$\times \frac{C}{4}$ and output it for feature concatenation along the channel dimension.


\noindent{\bf Global self-attention} (GSA). As shown in Figure~\ref{fig:sa} (b), our GSA branch is similar to the LSA branch on the first three steps of linear projection, feature split, and window partition.
Since self-attention in large window size suffers from huge computational costs, we apply a dimension reduction on the Key matrix $\bm{K}$ and Value matrix $\bm{V}$ after the window partition step of our GSA branch, as suggested in~\cite{linformer,wang2021pvtv1}.
The window size of $\bm{K}$ and $\bm{V}$ is reduced from $m \times m$ to $d \times d$ ($d<m$) by max-pooling, with proper reshape operations on the window dimensions.
The dimension-reduced matrices $\tilde{\bm{K}}$ and $\tilde{\bm{V}}$ are used to perform self-attention with the Query matrix $\bm{Q}$.
Finally, the GSA branch reshapes 
the attention feature
into $(H \times W) \times \frac{C}{4}$ and output it for feature concatenation.

\begin{figure}[t]
\vspace{-5mm}
	\centering
     \begin{overpic}[width=0.9\linewidth]
          {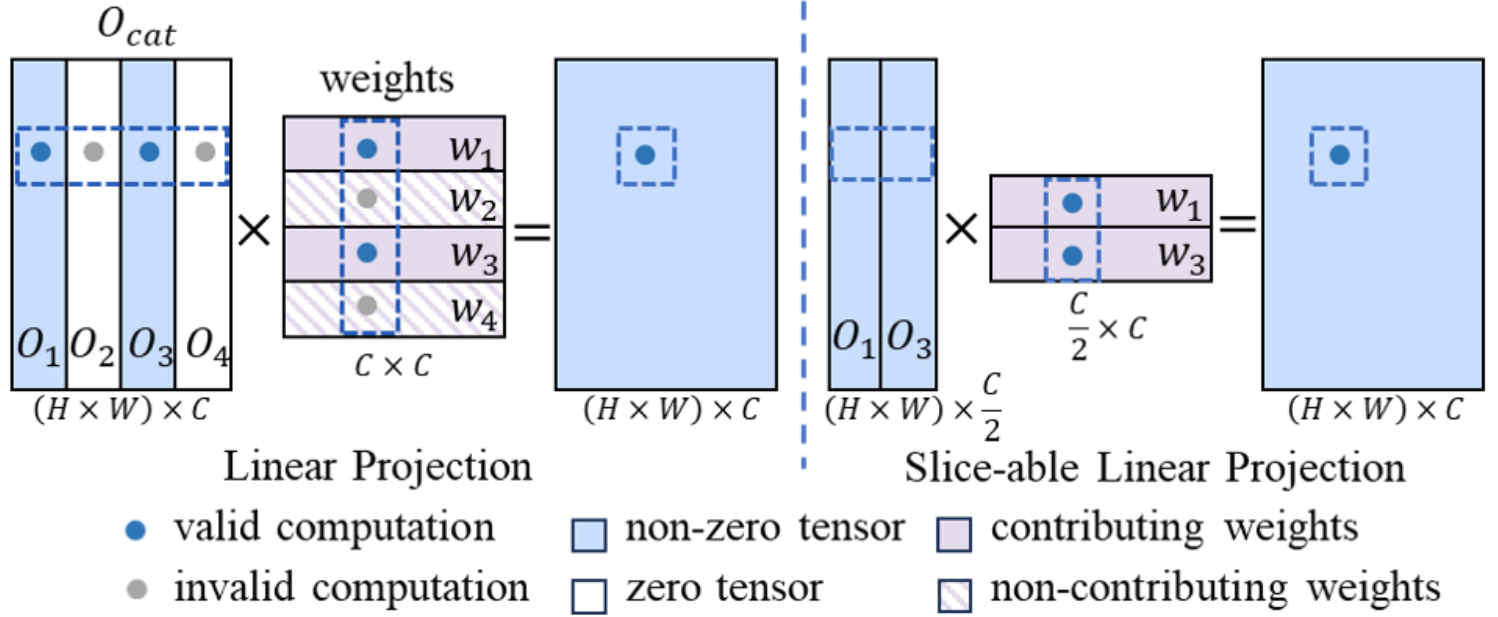}
     \end{overpic}
   \caption{Illustration of our slice-able linear projection. }
 \label{fig:slice}
  \vspace{4.2mm}
\end{figure}  

\noindent{\bf Slice-able linear projection}.
The output concatenated feature $\bm{O}_{cat}$ will be fused by linear projection in vanilla self-attention~\cite{2021swin_transformer}.
As shown in Figure~\ref{fig:slice}, denoting $\mathbf{W}\in\mathbb{R}^{C\times C}$ as the weight matrix of linear projection, we split it along the row dimension and get four sub-matrices of $\bm{W}_1$, $\bm{W}_2$, $\bm{W}_3$, and $\bm{W}_4$, all of size $\frac{C}{4}\times C$.
The vanilla linear projection is  equivalent to multiplying the feature matrix $\bm{O}_j$ with the corresponding weight matrix $\bm{W}_j$ for 
$j\in\{1,2,3,4\}$.
In our MSTB, 
if the  $j$-th
branch is bypassed, its output split feature $\bm{O}_j\in\mathbb{R}^{(H\times W)\times C/4}$ will be a comfortable zero tensor, and the corresponding matrix multiplication in linear projection also outputs a zero tensor and hence can be bypassed.

To save possible computational costs, we design a slice-able linear projection by removing the zero tensors in the output feature $\bm{O}_{cat}$ and the corresponding sub-matrices in the weight matrix $\bm{W}$.
In our slice-able version, we multiply the outcome split feature $\bm{O}_j$ and the corresponding weight sub-matrix $\bm{W}_j$ with a routing value of $\bm{r}^i_j=1$ for $j=$$1,2,3,4$, and denote them as $\bm{O}_j(r^i_j=1)$ and $\bm{W}_j(r^i_j=1)$, respectively.
Thus, the vanilla linear projection in our MSTB can be equally computed as
\vspace{-1.5mm}
\begin{equation}
\label{eqn:slicelp}
\bm{O}_{cat}\times \bm{W}
=
\left[\bm{O}_j(r^i_j=1)\right]\times\left[\bm{W}_j^{\top}(r^i_j=1)\right]^{\top}.
\end{equation}
\vspace{0mm}
The proposed slice-able linear projection reduces the computational complexity of the vanilla linear projection from $\mathcal{O}(HWC^{2})$ to $\mathcal{O}(\frac{1}{4}\sum_{j=1}^4{r^i_j}HWC^2)$.
Figure~\ref{fig:slice} gives an example of $r^i_2=r^i_4=0$, where $\bm{O}_{cat}\times\bm{W}=\left[\bm{O}_1, \bm{O}_3\right]\times\left[\bm{W}_1^{\top}, \bm{W}_3^{\top}\right]^{\top}$.

\subsubsection{Task-Aware Routing Controller} 
\label{sec:tarc}
The goal of our Task-Aware Routing Controller (TARC) is to predict the inference path of the feature extraction backbone for each ASSR task, consisting of an LR image and an SR scale.
As shown in Figure~\ref{fig:overall} (c), our TARC is a two-branch module to process the LR image and SR scale, respectively.
The image branch estimates a sampling probability vector $\bm{e}\in \mathbb{R}^{4N}$ for the $4N$ branches from the LR image, while the scale branch refines the probability vector by predicting an intensity scalar $\beta$ to indicate the difficulty of ASSR on this SR scale.

For the image branch, we estimate the sampling probability vector $\bm{e}$ from the LR image $\bm{I}^{LR}$ through two $3\times3$ convolutions followed by an average pooling and a linear projection.
For $i\in\{1,..,N\}$ and $j\in\{1,2,3,4\}$, the element $e_{j}^{i}$ of probability vector $\bm{e}$ is the probability of whether using the $j$-th self-attention branch of MSTBs in the $i$-th MSTG, estimated from the LR image $\bm{I}^{LR}$.
Therefore, the probability vector $\bm{e}$ varies for different LR images, which makes our TARC image-aware.

To further make our TARC module aware of SR scales (\ie, scale-aware), its scale branch transforms the SR scale $s$ to a scale-aware intensity scalar $\beta$ via three linear layers, as shown in Figure~\ref{fig:overall} (c).
Then the scale-aware intensity scalar $\beta$ is used to refine the probability vector $\bm{e}$ to output the task-aware probability vector $\bm{p}$:
\vspace{-1mm}
\begin{equation}
\label{eq:p_vector}
p^i_j =\min \left(\beta \times 4N \times \sigma\left(\bm{e}^i_j\right) / \sum_{i=1}^N\sum_{j=1}^4 \sigma\left(\bm{e}^i_j\right), 1\right), 
\end{equation}
where $\sigma$ is the sigmoid function.
%
%
We interpret $\beta$ as the intensity of our TARC to modulate all the $4N$ elements of the probability vector $\bm{e}$.
A small (or large) $\beta$ implies that our TARC tends to decrease (or increase) the element values of the task-aware probability vector $\bm{p}$.
%
%
%
%

\begin{table*}[th]
\centering
\footnotesize
\renewcommand{\arraystretch}{1.2}
\setlength\tabcolsep{3pt}
\caption{\small 
\textbf{Quantitative (PSNR (dB)) comparison of different feature extractors working with any arbitrary-scale upsampler on the DIV2K validation set}.
$\dagger$ indicates our implementation, while the others are directly evaluated with the released pre-trained models.
``-'' indicates unavailable results due to out-of-memory (OOM) issue. 
%
The best results are highlighted in \textbf{bold}. 
}
\vspace{3mm}
\resizebox{0.72\linewidth}{!}{
\begin{NiceTabular}{r|r|ccc|ccccc}
\hline
\multicolumn{2}{c}{Method}
 &   \multicolumn{3}{c|}{ In-scale } & \multicolumn{5}{c}{ Out-of-scale } \\
  \cline{1-10}
 {Upsampler}&{Feature Extractor} & $\times 2$ & $\times 3$ & $\times 4$ & $\times 6$ & $\times 12$ & $\times 18$ & $\times 24$ & $\times 30$ \\

 \hhline{-|-|---|-----}
\multirow{9}{*}{MetaSR~\cite{hu2019meta}}& EDSR-baseline~\cite{2017edsr}& 34.64 &30.93 &28.92 &26.61& 23.55& 22.03& 21.06& 20.37\\
&RDN~\cite{zhang2018rdn} & 35.00 &31.27 &29.25& 26.88& 23.73& 22.18& 21.17& 20.47\\
&RCAN$^{\dagger}$~\cite{zhang2018rcan}& 35.02 &31.29 &29.26 &26.89 & 23.74 &	22.20 &	21.18 &	20.48\\

&NLSA$^{\dagger}$~\cite{NLSN_Mei_2021_CVPR}&  -
&31.32 &	29.30 & 26.93 &	23.80 &	22.26 &	21.26 &	20.54

\\
&SwinIR~\cite{liang2021swinir}&  35.15 & 31.40& 29.33& 26.94 &23.80& 22.26& 21.26 &20.54\\	
& CAT-R-2$^{\dagger}$~\cite{cat_chen2022cross}&35.15 & 31.38 &29.29  &26.90 &23.77 
&22.23 & 21.24 &20.52
\\
\cline{2-10}
& Baseline (Ours)& 35.15 & 31.38 &	29.31 & 26.92 &	23.76 &	22.21	&21.20 &20.50\\
&TADT (Ours)&\textbf{35.21} &	\textbf{31.47} &	\textbf{29.41} & \textbf{27.02} &	\textbf{23.87} & \textbf{22.31}&	\textbf{21.31} & \textbf{20.58}\\


\hhline{-|-|--------}
\multirow{9}{*}{LIIF~\cite{2020liif}}& EDSR-baseline \cite{2017edsr} &  34.67 & 30.96 & 29.00 & 26.75 & 23.71 & 22.17 & 21.18 & 20.48 \\
& RDN\cite{zhang2018rdn} & 34.99 & 31.26 & 29.27 & 26.99 & 23.89 & 22.34 & 21.34 & 20.59 \\
&RCAN$^{\dagger}$~\cite{zhang2018rcan}& 35.02 &	31.30 &	29.31 & 27.02 &	23.91 &	22.36 &	21.33&	20.60
\\
&NLSA$^{\dagger}$~\cite{NLSN_Mei_2021_CVPR} & -

& 31.39 &	29.40 & 	27.11 &	23.98 &	22.41 &	21.38 &	20.64

\\
& SwinIR \cite{liang2021swinir}&  35.17 & 31.46 & 29.46 & 27.15 & 24.02 & 22.43 & 21.40 & 20.67 \\
& CAT-R-2$^{\dagger}$~\cite{cat_chen2022cross} &35.23 &31.49 &	29.49 
&27.18 &24.03&	22.45 &21.41 &	20.67\\
\cline{2-10}
&Baseline (Ours)  & 35.24  &31.51 &29.50 & 27.19 & 
24.04 &	22.46 &	21.42 &	20.69
\\
&TADT (Ours)& \textbf{35.28} & \textbf{31.55} &	\textbf{29.54} 
 &\textbf{27.23} &\textbf{24.07} &	\textbf{22.49} &	\textbf{21.45} &	\textbf{20.71}\\
\hhline{-|-|--------}
\multirow{9}{*}{LTE~\cite{lte-jaewon-lee}}&EDSR-baseline~\cite{2017edsr}&34.72& 31.02& 29.04 &26.81& 23.78& 22.23& 21.24 &20.53\\
&RDN~\cite{zhang2018rdn} &35.04 &31.32& 29.33& 27.04& 23.95& 22.40& 21.36& 20.64\\
&RCAN$^{\dagger}$~\cite{zhang2018rcan}& 35.02 &	31.30 &	29.31 & 		27.04 &	23.95 &	22.40 &	21.38 &	20.65
\\
&NLSA$^{\dagger}$~\cite{NLSN_Mei_2021_CVPR}& - &31.44 & 29.44 
&27.14 &24.03 &22.48 &	21.44 &	20.70
\\
&SwinIR~\cite{liang2021swinir} &35.24 & 31.50 &29.51& 27.20	&24.09&22.50 &21.47	&20.73	\\		
& CAT-R-2$^{\dagger}$ ~\cite{cat_chen2022cross}&35.27 &31.52 &29.52  &27.21 & 24.09 &	22.51&	21.46 &	20.73
\\
\cline{2-10}
&Baseline (Ours) & 35.27	& 31.53	& 29.52  & 27.21 & 24.08	& 22.50 &21.46	& 20.73\\
&TADT (Ours) & \textbf{35.31} &	\textbf{31.56} &	\textbf{29.56} &	\textbf{27.24} &	\textbf{24.10} & \textbf{22.52} &	\textbf{21.48} & \textbf{20.75}
\\
\hline
\end{NiceTabular}
}
\label{table:div2k}
 \vspace{-0.5mm}
\end{table*}

\begin{table*}[!t]
\centering
\footnotesize
\renewcommand{\arraystretch}{1.2}
\setlength\tabcolsep{3pt}
\caption{\small 
 \textbf{
Quantitative (PSNR (dB)) comparison of different ASSR methods on benchmark datasets
}.
$\dagger$ indicates our implementation, while the others are directly evaluated with the released pre-trained models.
The best results are highlighted in \textbf{bold}.
}
\vspace{4mm}
\resizebox{\linewidth}{!}{
\begin{NiceTabular}{r|r|ccccc|ccccc|ccccc}
\hline
\multicolumn{2}{c}{Method}
& 
\multicolumn{5}{c}{ B100 } & \multicolumn{5}{c}{ Urban100 } & \multicolumn{5}{c}{Manga109}\\
\cline{1-17}
 {Upsampler}&{Feature Extractor} &$\times 2$ & $\times 3$ & $\times 4$ & $\times 6$ & $\times 8$ 
  & $\times 2$ & $\times 3$ & $\times 4$ & $\times 6$ & $\times 8$
  & $\times 2$ & $\times 3$ & $\times 4$ & $\times 6$ & $\times 8$ \\
 \hline
\multirow{8}{*}{MetaSR~\cite{hu2019meta}}
&RDN~\cite{zhang2018rdn} 
&32.33& 29.26& 27.71& 25.90& 24.83 
&32.92& 28.82& 26.55& 23.99 &22.59 
&- &- &- &- &-\\
&RCAN$^{\dagger}$~\cite{zhang2018rcan} 
&  32.35 &29.29 &	27.73 & 25.91 & 24.83
 & 33.14 &	28.98 &	26.66 & 24.06 & 22.65
 & 39.37 &	34.44 &	31.26 & 26.97 & 24.5

\\
& NLSA${\dagger}$~\cite{NLSN_Mei_2021_CVPR}
& 32.35 &	29.30 & 	27.77 & 25.95 & 24.88
&33.25 &	29.12 &	26.80 & 24.20 & 22.78
&39.43 &	34.55 &	31.42 & 27.11 & 24.71
\\
& SwinIR~\cite{liang2021swinir} & 32.39 &29.31& 27.75& 25.94 &24.87 
&33.29 &29.12& 26.76& 24.16 & 22.75
&39.42&	34.58 &	31.34 & 26.96 & 24.62
\\
& CAT-R-2$^{\dagger}$~\cite{cat_chen2022cross} 
&32.40 & 29.29 & 27.72 & 25.91 &24.85
&33.35 &	29.11 &	26.69 & 24.11 & 22.73 
&39.49 &	34.52 &	31.17 & 26.86 & 24.54

\\
\cline{2-17}
&Baseline (Ours)
& 32.40 &29.32 &27.74 & 25.92 & 24.85
&33.34 &29.12 &26.74 & 24.14 &22.74
& 39.47 &34.53 & 31.28 & 26.88  & 24.53 
\\
& TADT (Ours)
& \textbf{32.47} & \textbf{29.36} & \textbf{27.80} & \textbf{25.97} & \textbf{24.91}
&\textbf{33.50} &\textbf{29.32} &\textbf{26.96} & \textbf{24.32}& \textbf{22.91}
& 
\textbf{39.57} &	\textbf{34.76} &\textbf{31.59} & \textbf{27.20} & \textbf{24.79}
\\

\hhline{-|-|-----|-----|-----}
 \multirow{8}{*}{LIIF~\cite{2020liif}} & RDN~\cite{zhang2018rdn}
&32.32& 29.26& 27.74& 25.98 &24.91 
&32.87& 28.82& 26.68& 24.20 &22.79
&39.22 &34.14 &	31.15 & 27.30 &25.00
\\
&RCAN$^{\dagger}$~\cite{zhang2018rcan} 
 & 32.36 &	29.29 &	27.77 & 26.01 &24.95
 &33.17 &29.03 &	26.86& 24.35& 22.92
& 39.37 &34.34 &31.31  &27.37 & 25.05
\\
&NLSA$^{\dagger}$~\cite{NLSN_Mei_2021_CVPR}
& 32.39 &	29.35 & 	27.83 & 26.06 &24.99
&33.44 &	29.35 &	27.15 & 24.58 & 23.07
 & 39.58&34.67 &31.65 & 27.65 & 25.26
\\
& SwinIR~\cite{liang2021swinir} 
&32.39 &29.34 &27.84 &26.07 &25.01
& 33.36& 29.33& 27.15 & 24.59 &23.14
& 39.53 & 34.65 &	31.67 & 27.66 & 25.28\\
& CAT-R-2$^{\dagger}$~\cite{cat_chen2022cross}

& 32.44 & 	29.38& 	27.86 & 26.09 & 25.02
& 33.58 &29.44	&27.23 & 24.67 & 23.19
& 39.53 &	34.66 &	31.69 &27.72 &25.31
\\
\cline{2-17}
& Baseline (Ours)
&32.44 &29.38 & 27.85 & 26.08 &25.03
& 33.54 &29.49 &27.27 & 24.68 & 23.22 
& 39.63	&34.74 &31.77 &27.74 &25.34
\\
&TADT (Ours)
&\textbf{32.46} & 	\textbf{29.41} & 	\textbf{27.87} & \textbf{26.10} & \textbf{25.05}
&\textbf{33.65} &	\textbf{29.58}& \textbf{27.37} & \textbf{24.75} & \textbf{23.27} 
 & \textbf{39.68} &\textbf{34.79} &	\textbf{31.83} & \textbf{27.84} & \textbf{25.39}
\\
\hline
\multirow{8}{*}{LTE~\cite{lte-jaewon-lee}} &
RDN~\cite{zhang2018rdn} 
&32.36& 29.30& 27.77 & 26.01 &24.95
&33.04& 28.97& 26.81 & 24.28 &22.88
& 39.25 &34.28	&31.27 &27.46 &25.09\\
& RCAN$^{\dagger}$~\cite{zhang2018rcan}
&32.37 &29.31 &	27.77 & 26.01 &24.96
& 33.13 &29.04 &26.88 & 24.33 &22.92 
& 39.41 &34.39 &	31.30 & 27.44 & 25.09 \\
&NLSA$^{\dagger}$~\cite{NLSN_Mei_2021_CVPR}
&32.43 & 29.39 &	27.86 & 26.08 &25.02 
& 33.56 &29.43 &	27.25 &24.62 & 23.15
& 39.64 &34.69 &	31.66 & 27.83 & 25.37

\\
& SwinIR~\cite{liang2021swinir} 
 &32.44& 29.39& 27.86 &26.09 &25.03
 & 33.50& 29.41& 27.24 & 24.62 &23.17
 & 39.60 &	34.76 &	31.76  & 27.81 &25.39\\
& CAT-R-2$^{\dagger}$~\cite{cat_chen2022cross}
&32.47 &29.39	& 27.87 & 26.09 & 25.03
& 33.60 &29.48 &27.27 & 24.68 & 23.21
&39.61 &34.75 &	31.76 & 27.84 & 25.39
\\
\cline{2-17}
& Baseline (Ours)
&32.46 & 29.39 &27.86 & 26.09 &25.04
& 33.67&	29.51 &	27.33 & 24.67 & 23.23
& 39.66 &	34.77 &	31.77 &27.85 &25.39
\\
& TADT (Ours)
& \textbf{32.47} &	\textbf{29.41} & \textbf{27.88} & \textbf{26.11} & \textbf{25.05}
& \textbf{33.70} & \textbf{29.57} & \textbf{27.36} & \textbf{24.72} & \textbf{23.26} 
& \textbf{39.72} &	\textbf{34.86} &\textbf{31.85} & \textbf{27.93} & \textbf{25.47}
\\
 \hline
 \multicolumn{2}{r}{ArbSR (ICCV'2021)~\cite{wang2021arbsr}} & 32.39 & 29.32 & 27.76 & 25.74 &24.55& 33.14 & 28.98 & 26.68 & 32.70 &22.13& 39.37 & 34.55 & 31.36 & 26.18 & 23.58
 \\
 \multicolumn{2}{r}{LIRCAN (IJCAI'2023)~\cite{ijcai2023p63}} & 32.42 & 29.36 &  27.82& - & - & 33.13 &  29.11 & 26.88& - & - &39.56 & 34.77 &  31.71 & - & -\\
 \multicolumn{2}{r}{EQSR (CVPR'2023)~\cite{EQSR}} & 32.46 & \textbf{29.42} & 27.86 & 26.07 & - & 33.62 & 29.53 & 27.30 & 24.66 & - & 39.44 & \textbf{34.89} & \textbf{31.86} & \textbf{27.97} & -
 \\
 \hline
\end{NiceTabular} 
}
\label{table:benchmark}
 \vspace{-2.5mm}
\end{table*}

With the scale-aware probability vector $\bm{p}$, each element $r^i_j\in\{0, 1\}$ ($i\in\{1,...,N\}$,\ $j\in\{1,2,3,4\}$) of the routing vector $\bm{r}$ can be drawn from Bernoulli sampling of $p^i_j$.
Since Bernoulli sampling is a non-differentiable operation, the gradient of the loss function $\mathcal{L}$ (will be introduced in \S\ref{sec:loss}) with respect to the routing value $r^i_j$ cannot be computed in backward pass.
To resolve this issue, as suggested in~\cite{zhou2016dorefa,2022DMVFN,huang2024safa}, we combine Straight-Through Estimator (STE)~\cite{bengio2013estimating,hubara2017quantized} with the Bernoulli sampling to make our TARC trainable.
The STE enables the backward pass of Bernoulli sampling to approximate the outgoing gradient by the incoming one.
Thus, we formalize the forward and backward passes of STE as:
\vspace{-1mm}
\begin{equation}
\label{eq:forback}
\begin{split}
\text{STE Forward Pass:}&
\ 
r^i_j\sim\operatorname{Bernoulli}\left(p^i_j\right),
\\
\text{STE Backward Pass:}&
\ 
\frac{\partial \mathcal{L}}{\partial r^i_j} =  \frac{\partial \mathcal{L}}{\partial p^i_j}.
\end{split}
\end{equation}
In this way, the Bernoulli sampling can be learnable by approximating the gradient $\partial\mathcal{L}/{\partial r^i_j}$ by the gradient $\partial\mathcal{L}/{\partial p^i_j}$.

\subsection{Loss Function}
\label{sec:loss}

The loss function $\mathcal{L}$ is a combination of the commonly used $\mathcal{L}_1$ loss and our newly proposed penalty loss $\mathcal{L}_{\beta}$ (which will be defined later) on the scale-aware intensity scalar $\beta$:
\vspace{-1mm}
\begin{equation}
\label{eq:loss}
\mathcal{L}=\mathcal{L}_1 + 
\lambda\mathcal{L}_{\beta},
\end{equation}
where $\lambda=2\times 10^{-4}$ is used to balance the two losses.
%
%
Here, penalty loss $\mathcal{L}_{\beta}$ is responsible to control scale-aware intensity $\beta$ in in Eqn.~(\ref{eq:p_vector}).
Since the scale-aware scalar $\beta$ implies the intensity of our TARC to select the $4N$ self-attention branches, it should be penalized by a loss function to constraint the computational budget.
A naive design is $\mathcal{L}_{\beta}=\beta$, but it potentially results in a small $\beta$ for all scale factors.
To avoid this problem, we simply incorporate a binary mask $M\in\{0,1\}$ on $\beta$ in $\lambda\mathcal{L}_{\beta}$, and $M$ is thresholded by the scale $s$ as follows:
\vspace{-1mm}
\begin{equation}
\label{eq:mask}
M\triangleq\beta\ge(\alpha_1 + \alpha_2 s^{\alpha_3}). 
\end{equation}
Then we can set the penalty loss $\mathcal{L}_{\beta}$ of the scalar $\beta$ by:
\vspace{-1mm}
\begin{equation}
\mathcal{L}_{\beta} = \beta M.
\end{equation}
\vspace{-2mm}
%
%
We set $\alpha_1=0.25$, $\alpha_2=0.25$, and $\alpha_3=0.5$.






\section{Experiments}
\label{sec:experiments}

\subsection{Experimental Setup}
\label{sec:general}

\noindent{\bf Dataset}.
Following previous ASSR works \cite{2020liif,lte-jaewon-lee,yang2021implicit,Chen_2023_CVPR}, we use the training set of DIV2K~\cite{Agustsson_2017_CVPR_Workshops} 
for model training.
For model evaluation, we report Peak Signal-to-Noise Ratio (PSNR) results on the DIV2K validation set~\cite{Agustsson_2017_CVPR_Workshops} and 
 benchmark datasets, including  B100~\cite{b100}, Urban100~\cite{Huang-CVPR-2015} and Manga109~\cite{manga109_mtap_matsui_2017}. 

\noindent{\bf Implementation details}.
We implement two variants of  TADT feature extractor: 1) the Baseline, \ie, the feature extraction backbone of our TADT (without the TARC), 2) the TADT.
%
We combine our TADT variants with the arbitrary scale upsamplers of MetaSR~\cite{hu2019meta}, LIIF~\cite{2020liif}, or LTE~\cite{lte-jaewon-lee} as our ASSR networks.
All the three TADT variants comprise $N=8$ MSTGs with $C=224$ channels.
Each MSTB in MSTGs has a global window size of $m=48$ and local window sizes of $\{m_1=4, m_2=8, m_3=16\}$.
Following~\cite{2020liif}, we set the channel dimension of the final output feature as $C_{out}=64$.

For network training, we employ the same experimental setup of previous works~\cite{2020liif,lte-jaewon-lee}.
To synthesize paired HR and LR data, given the images from the DIV2K training set and an SR scale $s$ evenly sampled from the uniform distribution $\mathcal{U}(1,4)$, we first crop $48s\times48s$ patches from the images as the ground-truth (GT) HR images, and then utilize bicubic downsampling 
to get the paired LR images of size $48 \times 48$.
We sample $48\times48$ pixels from the same coordinates of the SR image and the GT HR images to compute the training loss.

We train our TADT variants with each arbitrary-scale upsampler, \ie, MetaSR~\cite{hu2019meta}, LIIF~\cite{2020liif}, or LTE~\cite{lte-jaewon-lee}, as our ASSR networks described in \S\ref{sec:overview}.
%
%
Note that our Baseline is scale-agonostic and thus trained with only the $\mathcal{L}_{1}$ loss function by setting $\lambda=0$ in the loss function (\ref{eq:loss}).
Our TADT is trained on the pre-trained Baseline under the same settings, but with the loss function (\ref{eq:loss}).

\begin{table}[t]
\vspace{-2mm}
\centering
\footnotesize
\renewcommand{\arraystretch}{1.2}
\setlength\tabcolsep{3pt}
\caption{\small 
\textbf{Parameter amounts (M) and FLOPs (G)
of different feature extractors working with LIIF~\cite{2020liif}}, for ASSR at scale $s=2$, $3$, or $4$ on the DIV2K validation set.
``-'': the result is unavailable due to out-of-memory.
}
\vspace{3mm}
\resizebox{0.9\linewidth}{!}{
\begin{tabular}{r|r|rrr}
\toprule
 \multirow{2}{*}{Feature Extractor}&  \multirow{2}{*}{Params (M)} & \multicolumn{3}{c}{FLOPs (G)}  \\
&  &\multicolumn{1}{c}{$\times 2$} & \multicolumn{1}{c}{$\times 3$} & \multicolumn{1}{c}{$\times 4$} \\
 \midrule
RDN~\cite{zhang2018rdn} & 21.97 & 15567.48 & 6918.88 & 3891.87\\
RCAN~\cite{zhang2018rcan}& 15.33& 10774.88 & 4788.83 & 2693.72\\
SwinIR~\cite{liang2021swinir}& 11.60 &8832.28 &3923.36 &2227.08 \\
NLSA~\cite{NLSN_Mei_2021_CVPR} &39.58 & \multicolumn{1}{c}{-} & 13357.80 & 7513.77	\\
CAT-R-2~\cite{cat_chen2022cross} &11.63  & 8760.82 & 4038.19 & 2274.76\\
Baseline (Ours)  & 9.17& 7454.65 & 3407.59 & 1952.41\\
TADT (Ours)& 9.18  &6986.92  &3207.16 &1845.57 \\
\bottomrule
\end{tabular}
}
\label{table:cost}

     \vspace{-3mm}
\end{table}
\begin{figure*}[th]
    \centering
\vspace{-10mm}
\begin{overpic}
    [width=\linewidth]{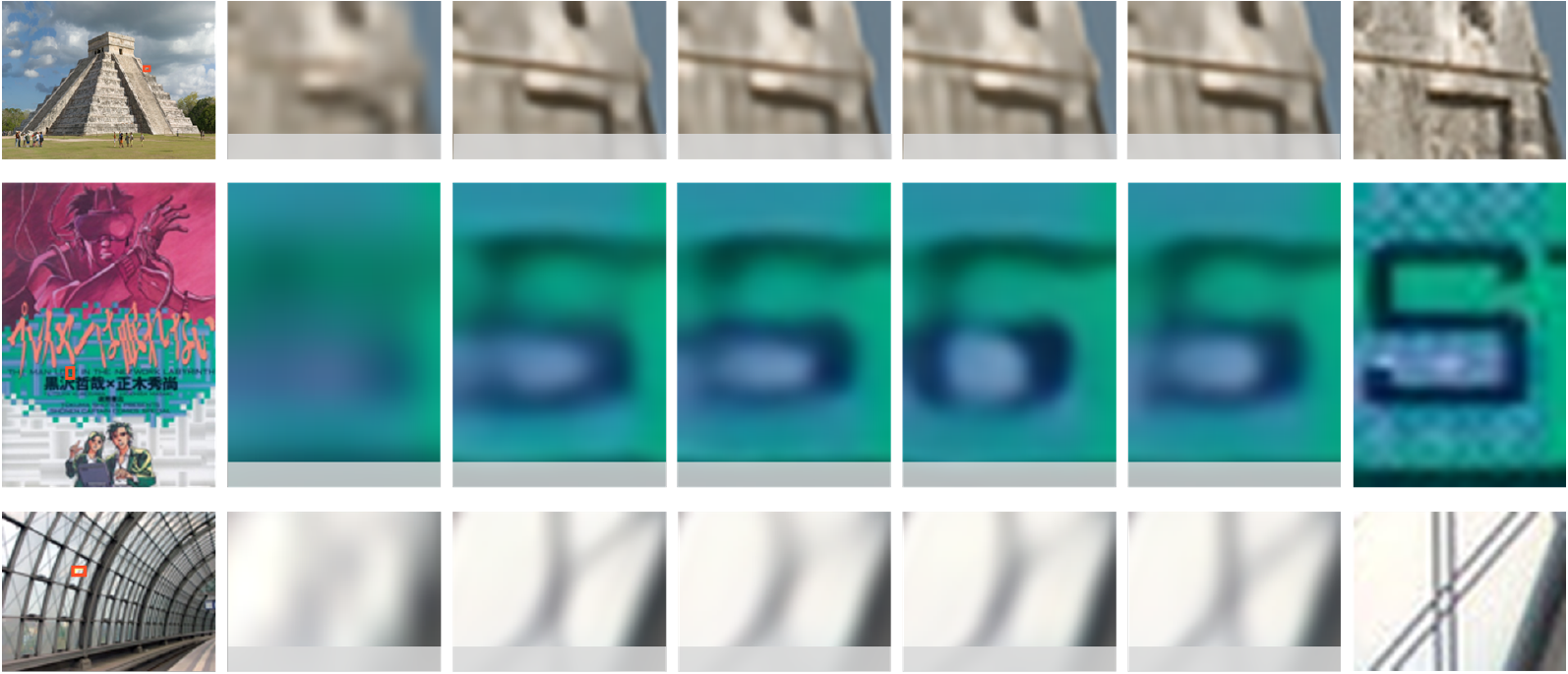}
    \put(-1.4,33){\rotatebox{90}{\scriptsize MetaSR~\cite{hu2019meta}}}
    \put(-1.4,18){\rotatebox{90}{\scriptsize LIIF~\cite{2020liif}}}
    \put(-1.4,2){\rotatebox{90}{\scriptsize LTE~\cite{lte-jaewon-lee}}}
    \put(3,31.7){\scriptsize DIV2K, $s=3$}
    \put(20,31.7){\scriptsize LR}
     \put(17.1,33.3){\scriptsize PSNR / FLOPs}
     \put(28.91,33.3){\scriptsize 28.02 dB / 12111.11 G}
    \put(33,31.7){\scriptsize NLSA~\cite{NLSN_Mei_2021_CVPR}}
    \put(44,33.3){\scriptsize 28.05 dB / 3568.53 G}
    \put(46,31.7){\scriptsize SwinIR~\cite{liang2021swinir}}
    \put(58.5,33.3){\scriptsize 28.07 dB / 3695.97 G}
    \put(61,31.7){\scriptsize CAT-R-2~\cite{cat_chen2022cross}}
    \put(73, 33.3){\scriptsize \textbf{28.21} dB / \textbf{2803.89} G}
    \put(75, 31.7){\scriptsize TADT (Ours)}
    \put(92.5,31.7){\scriptsize GT}
    \put(2,10.7){\scriptsize Manga109, $s=6$}
    \put(20,10.7){\scriptsize LR}
      \put(17.1,12.4){\scriptsize PSNR / FLOPs}
      \put(30,12.4){\scriptsize 26.51 dB / 920.94 G}
    \put(33,10.7){\scriptsize NLSA~\cite{NLSN_Mei_2021_CVPR}}
 
    \put(44,12.4){\scriptsize 26.48 dB / 291.50 G}
    \put(46,10.7){\scriptsize SwinIR~\cite{liang2021swinir}}

 \put(58.5, 12.4){\scriptsize 26.51 dB / 305.61 G}
\put(61,10.7){\scriptsize CAT-R-2~\cite{cat_chen2022cross}}

\put(73, 12.4){\scriptsize \textbf{26.59} dB / \textbf{254.08} G}
\put(75, 10.6){\scriptsize TADT (Ours)}
\put(92.5,10.7){\scriptsize GT}

\put(2,-1){\scriptsize Urban100, $s=12$}
    \put(20,-1){\scriptsize LR}
  \put(17.1,0.5){\scriptsize PSNR / FLOPs}

\put(30,0.5){\scriptsize  22.49 dB / 521.01 G}
\put(33,-1){\scriptsize NLSA~\cite{NLSN_Mei_2021_CVPR}}

\put(44,0.5){\scriptsize  22.47 dB / 152.09 G}
\put(46,-1){\scriptsize SwinIR~\cite{liang2021swinir}}

\put(58.8, 0.5){\scriptsize  22.44 dB / 152.81 G}
\put(61,-1){\scriptsize CAT-R-2~\cite{cat_chen2022cross}}

\put(73,0.5){\scriptsize  \textbf{22.55} dB / \textbf{120.79} G}
\put(75, -1){\scriptsize TADT (Ours)}
\put(92.5,-1){\scriptsize GT}
\end{overpic}
\vspace{-1mm}
    \caption{
 {\bfseries Visual comparison of different ASSR networks for natural image ASSR}.
 The ASSR networks are made up of different feature extractors and arbitrary-scale upsamplers, \ie, MetaSR~\cite{hu2019meta} ($1$-st row), LIIF~\cite{2020liif} ($2$-nd row), and LTE~\cite{lte-jaewon-lee} ($3$-rd row).
 The highlighted regions are zoomed in for better view.
    } \label{fig:vis_comp1}
    \vspace{3mm}
\end{figure*}

\subsection{Main Results}
\noindent{\bf Quantitative results}.
We compare our TADT variants with six off-the-shelf feature extractors, \ie, EDSR-baseline~\cite{2017edsr}, RDN~\cite{zhang2018rdn}, RCAN~\cite{zhang2018rcan}, NLSA~\cite{NLSN_Mei_2021_CVPR}, SwinIR~\cite{liang2021swinir}, and CAT-R-2~\cite{cat_chen2022cross}. The PSNR results on the DIV2K validation set and the five benchmark datasets are summarized in Table \ref{table:div2k} and Table \ref{table:benchmark}, respectively.
We also provide results of other ASSR methods including ArbSR~\cite{wang2021arbsr}, LIRCAN~\cite{ijcai2023p63} and EQSR~\cite{EQSR} in Table \ref{table:benchmark} for reference.
Our TADT achieves overall superior performance across all the test sets and SR scales, when working with MetaSR~\cite{hu2019meta}, LIIF~\cite{2020liif}, or LTE~\cite{lte-jaewon-lee}. More results can be found in our supplementary materials.
%

\noindent{\bf Qualitative results.} We provide the qualitative results of TADT along with  comparison feature extractors  in Figure~\ref{fig:vis_comp1}.
Here, we compare our TADT with NLSA~\cite{NLSN_Mei_2021_CVPR},
SwinIR~\cite{liang2021swinir}, and CAT-R-2~\cite{cat_chen2022cross}, since they achieve comparable PSNR results in Tables~\ref{table:div2k} and~\ref{table:benchmark}.
We observe that the SR results of different upsamplers working with our TADT exhibit more accurate structures, \eg, the shape of character ``S'' (the $2$-nd row) and the shape of X-type steel pole (the $3$-rd row), as well as the textures of stone (the $1$-st row), than the SR results of these upsamplers working with the other feature extractors. 

\noindent
\textbf{Computational costs}.
In Table \ref{table:cost}, we summarize the parameter amounts and computational costs of different feature extractors when working with the upsampler LIIF~\cite{2020liif}.
One can see that our Baseline and  TADT  are more efficient on both aspects than other competitors.

\subsection{Ablation Study}
\label{sec:ablation}
Here, we perform ablation studies to investigate the working mechanism of our TADT feature extractor on image ASSR tasks.
In all experiments, we use LIIF~\cite{2020liif} as the arbitrary-scale upsampler to work with our TADT feature extractor.

\vspace{-1mm}
\begin{figure}[t]
    \centering
    \vspace{0mm}
 \begin{overpic}
     [width=0.65\linewidth]
{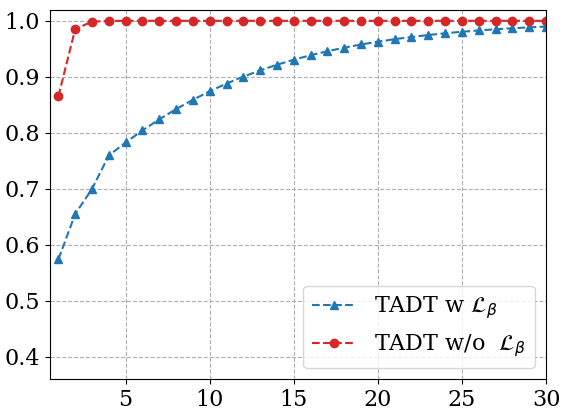}
\put(-5,40){\rotatebox{90}{\footnotesize $\beta$ }}
\put(45,-3){\footnotesize Scale $s$}
\end{overpic}
\vspace{2mm}
    \caption{ 
   {\bfseries The predicted $\beta$ in our TADT} \textsl{w.r.t.} different SR scales $s$.
  }
 \label{fig:beta}
    \vspace{4mm}
\end{figure}

\noindent{\bf 1) Does the scale branch in our TARC contribute to our TADT on scale-aware ASSR performance}?
To answer this question, we compare our TADT  with two other variants: a) directly using scale-agnostic $\beta=0.5$ and b) manually setting $\beta=0.25s$, where $s$ is the SR scale.
As summarized in Table~\ref{table:beta}, although achieving reasonable results on $\times 2$ upsampling, our ASSR network with $\beta=0.5$ in our TARC suffers from inferior PSNR results on upsampling for higher scales when compared with our TADT.
Manually setting $\beta=0.25s$ enables our ASSR network to achieve comparable results with our TADT at high SR scales of $s=6,8$, but falls short in ASSR at lower scales, \eg, 0.08 dB lower than our TADT on $\times 2$ SR tasks.
Our TADT well balances the performance across all the scales.
As revealed in Figure~\ref{fig:beta} $\beta$ in our TADT basically grows with the SR scale in ASSR, which is consistent with our intent on its role of intensity indicator.

\noindent{\bf 2) The influence of penalty loss $\mathcal{L}_{\beta}$ to our ASSR network}. 
We investigate this point by comparing our ASSR networks trained with or without using $\mathcal{L}_{\beta}$.
In Figure~\ref{fig:beta}, we visualize the curves of predicted $\beta$ \textsl{v.s.} SR scales after training our TADT based ASSR network with $\mathcal{L}_{\beta}$ and without $\mathcal{L}_{\beta}$.
%
%
We observe that,, without using $\mathcal{L}_{\beta}$ in training, our ASSR networks are prone to predict saturated $\beta$ when the SR scale increases, with higher computational costs. 
As summarized in Table~\ref{table:loss}, training our TADT without $\mathcal{L}_{\beta}$ obtains a minor PSNR increase of 0.02dB for $\times 2$ SR tasks on the DIV2K validation set, but also leads to a 388.75G FLOPs growth on computational costs.
Therefore, it is necessary to use our intensity penalty loss $\mathcal{L}_{\beta}$ in training our ASSR networks for computational efficiency.

\begin{table}[t]
\vspace{-2mm}
\centering
\caption{
\textbf{PSNR (dB) results of our ASSR network with different designs of intensity indicator $\beta$} on Urban100~\cite{Huang-CVPR-2015}. 
}
\vspace{4mm}
\resizebox{0.92\linewidth}{!}{
\begin{tabular}{c|ccc|cc}
\toprule
\multirow{2}{*}{$\beta$} & \multicolumn{3}{c|}{ In-scale } & \multicolumn{2}{c}{ Out-of-scale } \\
&$\times 2$ & $\times 3$ & $\times 4$ & $\times 6$ & $\times 8$ \\
 \midrule
 $\beta = 0.5$  
 &
33.66 &29.53 &27.33 &	24.72 &	23.23	

 \\
  $\beta = 0.25s $  &33.57	&29.53 &27.34 & 24.74 &	23.24	
 \\
%
Our TARC & 
33.65 & 29.58& 27.37 & 24.75 &	23.27
  \\
\bottomrule
\end{tabular}
}
 \vspace{-1mm}
\label{table:beta}
\end{table}
\begin{table}[t]
\vspace{-1mm}
\centering
\footnotesize
\renewcommand{\arraystretch}{1.2}
\setlength\tabcolsep{3pt}
\caption{\small
\textbf{
Results of PSNR (dB) and FLOPs (G) by our TADT  trained with (w) or without (w/o) the intensity loss $\mathcal{L}_{\beta}$} on the DIV2K validation set.
}
\vspace{4mm}
\resizebox{0.92\linewidth}{!}{
\begin{tabular}{c|c|c|c|c|c|c}
\toprule
 \multirow{2}{*}{Feature Extractor} &\multicolumn{2}{c|}{$\times 2$} & \multicolumn{2}{c|}{$\times 3$} &\multicolumn{2}{c}{$\times 4$} \\
\cline{2-7}
& PSNR & FLOPs & PSNR & FLOPs & PSNR & FLOPs \\
\midrule
TADT, w $\mathcal{L}_{\beta}$  &35.28& 6986.91& 31.55&3207.16& 29.54 &1845.57\\
TADT, w/o $\mathcal{L}_{\beta}$ &35.30& 7375.66	& 31.56 &3378.02&29.55 & 1936.01
 \\  
\bottomrule
\end{tabular}
}
\label{table:loss}
 \vspace{-1.8mm}
\end{table}


\section{Conclusion}
\label{sec:conlusion}
In this paper, we proposed an efficient feature extractor, \ie, the Task-Aware Dynamic Transformer (TADT), for image ASSR.
The proposed TADT contains cascaded multi-scale transformer groups (MSTGs) as the feature extraction backbone and a task-aware routing controller (TARC).
Each MSTG group consists of two multi-scale transformer blocks (MSTBs). Each MSTB block has three local self-attention branches to learn useful multi-scale representations and a global self-attention branch to extract distant correlations.
Given an inference task, \ie, an input image and an SR scale, our TARC routing controller predicts the inference paths  within the self-attention branches of our TADT backbone.
With task-aware dynamic architecture, our TADT achieved efficient ASSR performance when compared to the mainstream feature extractors.

\begin{ack}
This research is supported in part by The National Natural Science Foundation of China (No. 12226007 and 62176068) and  the Open Research Fund from the Guangdong Provincial Key Laboratory of Big Data Computing, The Chinese University of Hong Kong, Shenzhen, under Grant No. B10120210117-OF03. 
\end{ack}

\clearpage
\bibliography{ecai_arxiv}
\clearpage
\appendix

\section{Content}
\label{sec:cotent}
In this supplementary file, we further elaborate our Task-Aware Dynamic Transformer (TADT) as an efficient feature extractor for Arbitrary-Scale Image Super-Resolution.
Specifically, we present
\begin{itemize}
\vspace{-2mm}
   \item more ablation studies of our TADT in Se{sec:ablation};
   \item more quantitative results of our TADT in \cref{sec:results};
    \item more visual comparisons of our TADT and other feature extractors on natural image ASSR in \cref{sec:vis}.
    \vspace{-2mm}
\end{itemize}

\section{Ablation Studies}
\label{sec:ablation}
Here, we perform more ablation studies to investigate the working mechanism of our TADT feature extractor on image ASSR.
Similar to the main paper, all ablation experiments here are conducted on our TADT integrated with the arbitrary-scale upsampler LIIF~\cite{2020liif}.

\noindent{\bf 1) Effectiveness of using binary mask $\bm{M}$ in our intensity loss $\mathcal{L}_{\beta}$}.
To illustrate this point, we remove the binary mask $\bm{M}$ in $\mathcal{L}_{\beta}$ and directly set $\mathcal{L}_{\beta}=\beta$ in the loss function $\mathcal{L}$ to train our TADT.
We  visualize the $\beta$ with different $s$ in our TADT when $\mathcal{L}_{\beta}=\beta$  by the 
\textcolor[RGB]{214, 39, 40}{red}
curve in Figure~\ref{fig:beta}.
One can see that, the $\beta$ value in our TADT goes a slight ascent and then sweep down to about 0.1, which is unreasonable.
Quantitative results reported in Table~\ref{table:beta} further demonstrate that excluding the mask $\bm{M}$ from the intensity loss $\mathcal{L}_{\beta}$ leads to a significant performance drop, with a decrease of 0.09 dB for $\times 8$ upsampling.
This validates the effectiveness of using a binary mask $\bm{M}$ in our intensity loss $\mathcal{L}_{\beta}$ to train our TADT for image ASSR.

\begin{figure}[!h]
    \centering
    \vspace{-2mm}
 \includegraphics[width=0.78\linewidth]
{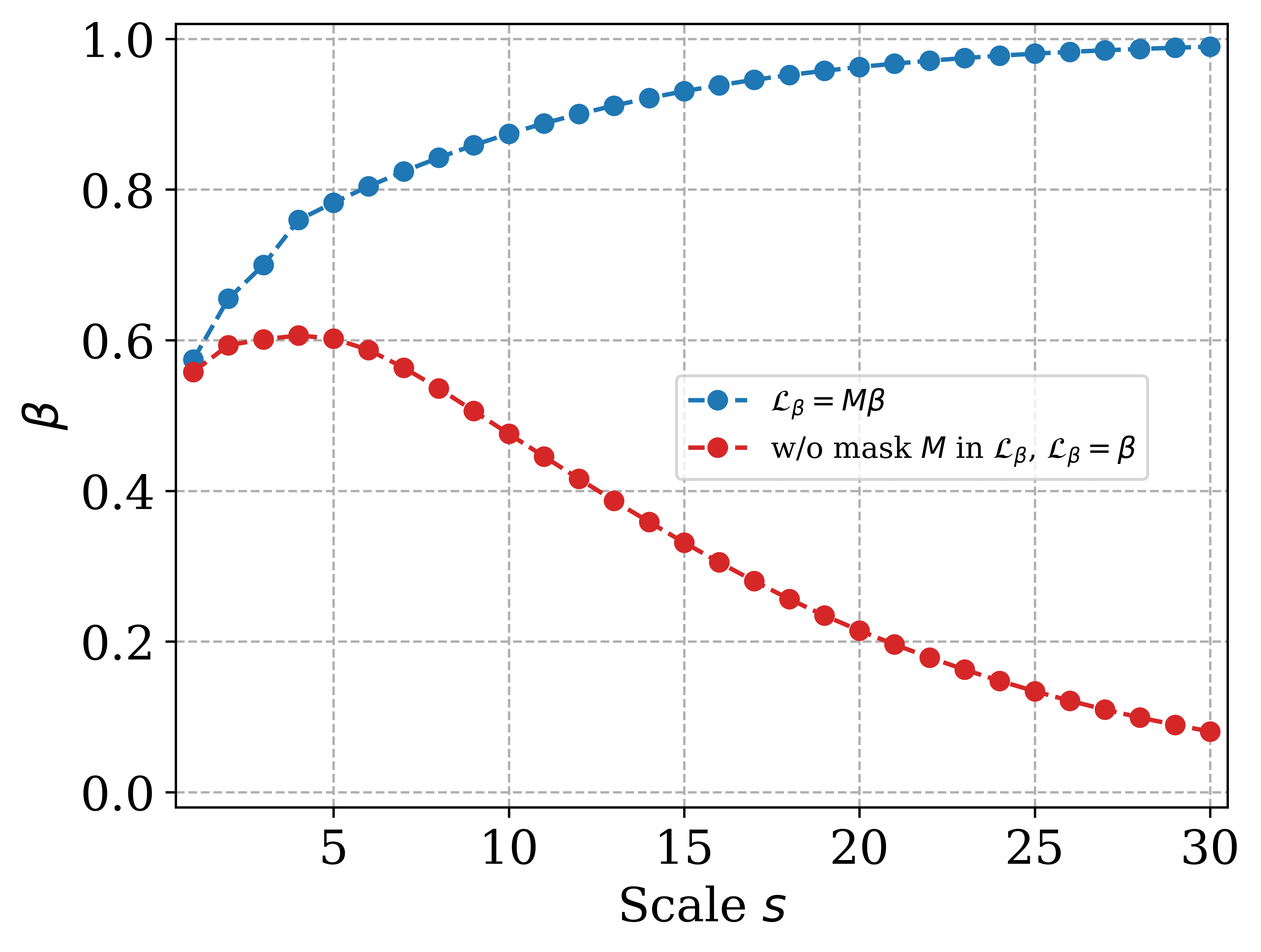}
\vspace{-1mm}
    \caption{\textbf{The predicted $\beta$ in our TADT} \textit{w.r.t.} different SR scales $s$.
  }
 \label{fig:beta}
    \vspace{4mm}
\end{figure}

\begin{table}[th]
\vspace{-0mm}
\centering
\caption{\textbf{PSNR (dB) results of our TADT  trained with $\mathcal{L}_{\beta}$ using the binary mask or not} on Urban100~\cite{Huang-CVPR-2015}.
 }
\vspace{3mm}
\resizebox{0.9\linewidth}{!}{
\begin{tabular}{c|ccc|cc}
\toprule
\multirow{2}{*}{$\mathcal{L}_{\beta}$} &\multicolumn{3}{c|}{ In-scale } & \multicolumn{2}{c}{ Out-of-scale } \\
 &$\times 2$ & $\times 3$ & $\times 4$ & $\times 6$ & $\times 8$ \\
 \midrule

 $\mathcal{L}_{\beta} = \beta $
&33.66 &	29.56 &27.36 &	24.73 &	23.18
\\
$\mathcal{L}_{\beta} = \beta M $ 
& 33.65 & 29.58& 27.37 & 24.75 &	23.27
\\
\bottomrule
\end{tabular}
}
\vspace{-1mm}
\label{table:beta}
\end{table}

\begin{table}[!t]
\vspace{-0mm}
\caption{
\textbf{PSNR (dB) results of our Baseline with (w) or without (w/o) global self-attention (GSA) branch} on the DIV2K validation set.
}
\vspace{3mm}
\centering
\resizebox{\linewidth}{!}{
\begin{tabular}{c|ccc|ccccc}
\toprule
 \multirow{2}{*}{Extrator} & \multicolumn{3}{c|}{ In-scale } & \multicolumn{2}{c}{ Out-of-scale } \\
&  $\times 2$ & $\times 3$ & $\times 4$ & $\times 6$ & $\times 12$  \\ 
\midrule
w GSA &
35.24 & 31.51 & 29.50  &27.19 &24.04  \\
w/o GSA
 &35.18	&31.45 &	29.45 &	27.14 &	24.02
\\
\bottomrule
\end{tabular}
}
\vspace{-1mm}
\label{table:global}
\end{table}

\begin{table}[!t]
\vspace{-0mm}
\caption{
\textbf{PSNR (dB) results of different dimension-reduction operations in global self-attention} on DIV2K validation set.
}
\vspace{3mm}
\centering
\resizebox{\linewidth}{!}{
\begin{tabular}{c|ccc|cc}
 \toprule
 \multirow{2}{*}{Dimension Reduction} & \multicolumn{3}{c|}{ In-scale } & \multicolumn{2}{c}{ Out-of-scale } \\
&  $\times 2$ & $\times 3$ & $\times 4$ & $\times 6$ & $\times 12$  \\
\midrule
 Random Matrix 
 &35.22 &	31.48 &	29.49 & 	27.16 &	24.00
\\
 Avgpooling
& 35.23 &31.49 & 29.49 & 27.18 &	24.04

 \\
 Maxpooling 
 &35.24 &31.51& 29.50& 27.19 &24.04\\
\bottomrule
\end{tabular}
}
\vspace{-1mm}
\label{table:proj}
\end{table}

\noindent{\bf 2) Importance of the global self-attention (GSA) branch in the MSTB of our feature extraction backbone}.
To study this aspect, we conduct experiments by evaluating our ASSR network with or without the GSA branch in each MSTB of our feature extraction backbone.
Here, we use the Baseline instead of our TADT to use the complete feature extraction backbone for fully comparison.
As shown in Table \ref{table:global}, our ASSR network using the Baseline with GSA achieves a performance gain of 0.06 dB on PSNR over that without GSA, on the DIV2K validation set for $\times 2$ SR.
This validates the importance of GSA branch in our feature extraction backbone for image ASSR.

\noindent{\bf 3) Investigation on dimension reduction in the GSA branch}.
To this end, we explore other dimension reduction variants, \eg, ``Random Matrix'' and ``Avgpooling'' for the GSA branch in our Baseline variant.
Here, ``Random Matrix'' performs dimension reduction by a linear projection matrix of size $m^2\times d^2$, which is randomly sampled from a normal distribution.
For ``Avgpooling'', we just replaces the ``maxpooling'' operation in the GSA by average pooling.
As shown in Table \ref{table:proj}, the ``Maxpooling'' employed in GSA achieves slightly better results (0.01$\sim$ 0.04dB) on ASSR tasks at most scales.
Thus, we use ``Maxpooling'' for dimension reduction in the GSA branch.

\noindent{\bf 4) {Investigation on the sensitivity of hyper-parameters.}} In Eqn~\ref{eq:mask}, 
we use $\alpha_1$, $\alpha_2$ and  $\alpha_3$ to penalize the predicted $\beta$ by a loss $\mathcal{L}_{\beta} = \beta M$, using a binary value $M\triangleq\beta\ge(\alpha_1 + \alpha_2 s^{\alpha_3})$ to avoid small $\beta$ values. $\alpha_1$ is the lower bound of $\beta$ to be penalized. $\alpha_2$ and $\alpha_3$ should be properly set to adjust the binary value $M$ according to the scale $s$. We report the experimental results achieved by different hyper-parameter settings on Urban100 in Table~\ref{table:hyper}. We observe that our TADT is not very sensitive to the values of  $\alpha_1$, $\alpha_2$ and  $\alpha_3$ once they are properly set, but our setting reported in paper achieves an overall better performance.

\begin{table}[!t]
\vspace{-0mm}
\caption{
\textbf{PSNR (dB) results of different hyper-parameter settings} on Urban100.
}
\vspace{3mm}
\centering
\resizebox{\linewidth}{!}{
\begin{tabular}{c|ccc|cc}
 \toprule
 \multirow{2}{*}{Hyper-parameters} & \multicolumn{3}{c|}{ In-scale } & \multicolumn{2}{c}{ Out-of-scale } \\
&  $\times 2$ & $\times 3$ & $\times 4$ & $\times 6$ & $\times 8$  \\
\midrule
 $\alpha_1=0.25,\alpha_2=0.25,\alpha_3=0.50$ &
 33.65  &29.58 & 27.37&  24.75 & 23.27
\\
$\alpha_1=0.00,\alpha_2=0.25,\alpha_3=1.00$ 
& 33.62 & 29.57 &  27.37 & 24.77 & 23.27
\\
$\alpha_1=0.00,\alpha_2=0.25,\alpha_3=0.50$ 
& 33.64 & 29.52 & 27.32 & 24.61 & 23.02
\\
$\alpha_1=0.25,\alpha_2=0.50,\alpha_3=0.50$ 
 &33.64  &29.56  &27.36 & 24.74  &23.25
\\
\bottomrule
\end{tabular}
}
\vspace{-1mm}
\label{table:hyper}
\end{table}

\vspace{-2mm}
\section{More Quantitative Results}
\label{sec:results}
In Table~\ref{table:benchmark}, we provide more quantitative results on Set5, Set14, 
B100, Urban100, and Manga109.
\vspace{-2mm}
\section{More Visual Comparison on Image ASSR}
\label{sec:vis}
\begin{figure*}[!h]
    \centering
    \vspace{-3mm}
\begin{overpic}
    [width=0.95\linewidth]{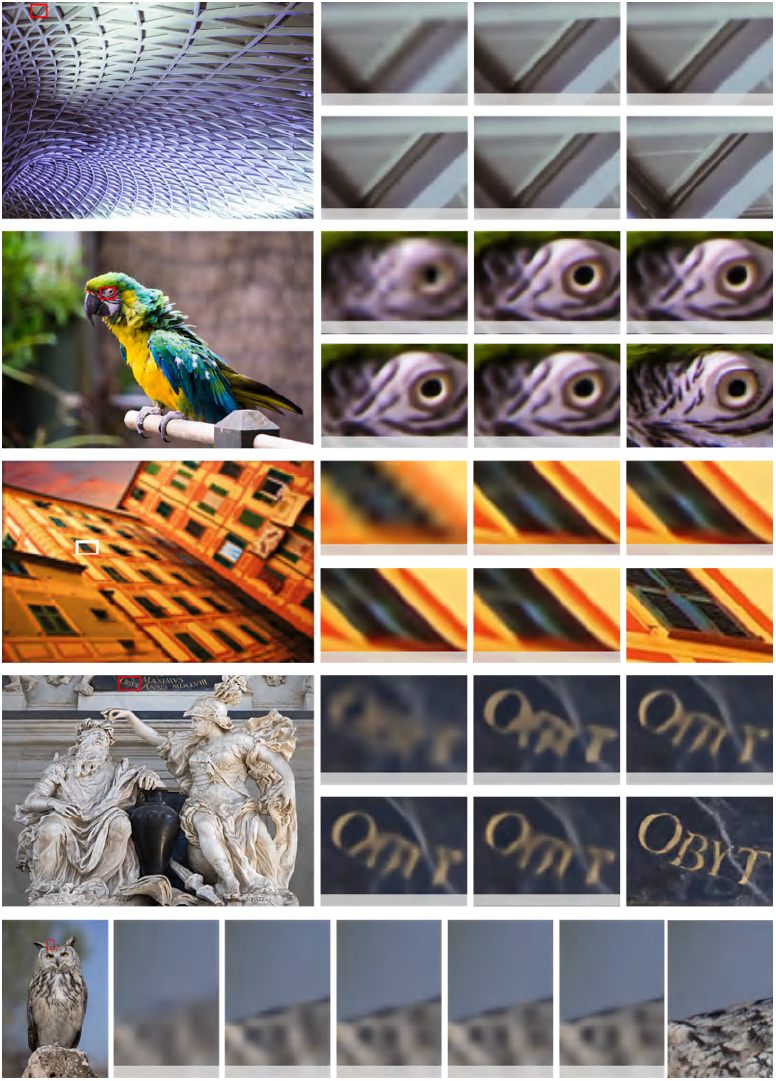}

  \put(10.5,78.95){\scriptsize Urban100, $s=6$}
 \put(35,89.5){\scriptsize LR}
     \put(33.5, 90.5){\scriptsize PSNR / FLOPs}

   \put(46.5,90.5){\scriptsize 28.21 dB /  3085.67 G}
 \put(49,89.5){\scriptsize NLSA~\cite{NLSN_Mei_2021_CVPR}}

 \put(60.5,90.5){\scriptsize 28.15 dB / 919.66 G}
 \put(62,89.5){\scriptsize SwinIR~\cite{liang2021swinir}}

 \put(32.3,80){\scriptsize 28.32 dB / 980.50 G}
 \put(35, 78.95){\scriptsize  CAT-R-2~\cite{cat_chen2022cross}}

 \put(46.5,80){\scriptsize 28.57 dB / 847.90 G}
 \put(48,78.95){\scriptsize TADT (Ours)}
\put(64,78.95){\scriptsize GT}

  \put(10.5,57.6){\scriptsize Urban100, $s=6$}
 \put(35,68.3){\scriptsize LR}
     \put(33.5, 69.4){\scriptsize PSNR / FLOPs}

   \put(46.5,69.4){\scriptsize 34.18 dB /  3258.52 G}
 \put(49,68.3){\scriptsize NLSA~\cite{NLSN_Mei_2021_CVPR}}

 \put(60.5,69.5){\scriptsize 34.20 dB / 987.78 G}
 \put(62,68.3){\scriptsize SwinIR~\cite{liang2021swinir}}

 \put(32.3,58.8){\scriptsize 34.18 dB / 1050.53 G}
 \put(35, 57.6){\scriptsize  CAT-R-2~\cite{cat_chen2022cross}}

 \put(46.5,58.8){\scriptsize 34.30 dB / 838.15 G}
 \put(48.5,57.6){\scriptsize TADT (Ours)}
\put(64,57.6){\scriptsize GT}
%
  \put(10.5,37.9){\scriptsize Urban100, $s=8$}
 \put(35,47.8){\scriptsize LR}
     \put(33.5, 48.8){\scriptsize PSNR / FLOPs}

   \put(46.5,48.8){\scriptsize 26.59 dB /  515.52 G}
 \put(49,47.8){\scriptsize NLSA~\cite{NLSN_Mei_2021_CVPR}}

 \put(60.5,48.8){\scriptsize 26.58 dB / 152.09 G}
 \put(62,47.8){\scriptsize SwinIR~\cite{liang2021swinir}}

 \put(32.3,39){\scriptsize 26.60 dB /  152.81 G}
 \put(35, 37.9){\scriptsize  CAT-R-2~\cite{cat_chen2022cross}}

 \put(46.5,39){\scriptsize 26.74 dB / 127.15 G}
 \put(48.5,37.9){\scriptsize TADT (Ours)}
\put(64,37.9){\scriptsize GT}

 \put(10.5,15.3){\scriptsize DIV2K, $s=12$}
 \put(35,26.5){\scriptsize LR}
     \put(33.5, 27.6){\scriptsize PSNR / FLOPs}

   \put(46.5,27.6){\scriptsize 24.67 dB /  1045.59 G}
 \put(49,26.5){\scriptsize NLSA~\cite{NLSN_Mei_2021_CVPR}}

 \put(60.5,27.6){\scriptsize 24.71 dB / 331.11 G}
 \put(62,26.5){\scriptsize SwinIR~\cite{liang2021swinir}}

 \put(32.3,16.4){\scriptsize 24.71 dB /  350.18 G}
 \put(35, 15.3){\scriptsize  CAT-R-2~\cite{cat_chen2022cross}}

 \put(46.5,16.4){\scriptsize 24.73 dB / 338.85 G}
 \put(48.5,15.3){\scriptsize TADT (Ours)}
\put(64,15.3){\scriptsize GT}
%
%
\put(1.5,-0.55){\scriptsize  DIV2K, $s=12$}
\put(15,-0.55){\scriptsize LR}
     \put(12.5,0.58){\scriptsize PSNR / FLOPs}
   \put(21.4,0.55){\scriptsize 26.32 dB / 814.83 G}
 \put(23.5,-0.55){\scriptsize NLSA~\cite{NLSN_Mei_2021_CVPR}}

 \put(32,0.58){\scriptsize  26.34 dB /  261.40 G}
 \put(34,-0.55){\scriptsize SwinIR~\cite{liang2021swinir}}

 \put(42, 0.58){\scriptsize 26.34 dB / 786.31 G}
 \put(44, -0.55){\scriptsize  CAT-R-2~\cite{cat_chen2022cross}}

 \put(52,0.58){\scriptsize 26.36 dB / 745.76 G}
 \put(53,-0.55){\scriptsize TADT (Ours)}
\put(65,-0.55){\scriptsize GT}
\end{overpic}
\vspace{3mm}
    \caption{ {
   \bfseries Visual comparison of feature extractors integrated with LTE~\cite{lte-jaewon-lee} on super-resolution for natural images at scale 6, 8, 12.}  The highlighted regions are zoomed in for better view.
    } \label{fig:vis_comp6}
    \vspace{-4mm}
\end{figure*}

In Figures~\ref{fig:vis_comp1}-\ref{fig:vis_comp6}, we provide more visual comparison results of different feature extractors working with three arbitrary-scale upsamplers, \ie, MetaSR~\cite{hu2019meta}, LIIF~\cite{2020liif}, and LTE~\cite{lte-jaewon-lee}, on the image ASSR task.

\begin{table*}[!t]
\centering
\footnotesize
\renewcommand{\arraystretch}{1.2}
\setlength\tabcolsep{3pt}
\caption{\small 
 \textbf{
Quantitative comparison of PSNR (dB) results by different feature extractors} working with arbitrary-scale upsamplers on five benchmark datasets.
$\dagger$ indicates our implementation, while the others are directly evaluated with the released pre-trained models.
The best results are highlighted in \textbf{bold}.
}
\vspace{4mm}
\resizebox{\linewidth}{!}{
\begin{NiceTabular}{r|r|ccc|ccc|ccc|ccc|ccc}
\hline
\multicolumn{2}{c}{Method}
& 
\multicolumn{3}{c}{ Set5 } & \multicolumn{3}{c}{ Set14 } & \multicolumn{3}{c}{ B100 } & \multicolumn{3}{c}{ Urban100 } & \multicolumn{3}{c}{Manga109}\\
\cline{1-17}
 {Upsampler} & {Feature Extractor} &
 $\times 2$ & $\times 3$ & $\times 4$ & $\times 2$ & $\times 3$ & $\times 4$  & $\times 2$ & $\times 3$ & $\times 4$ &  $\times 2$ & $\times 3$ & $\times 4$ & $\times 2$ & $\times 3$ & $\times 4$ \\
 \hline
\multirow{8}{*}{MetaSR~\cite{hu2019meta}}
&RDN~\cite{zhang2018rdn} &38.22 & 34.63& 32.38 &33.98& 30.54& 28.78&  32.33& 29.26& 27.71& 32.92& 28.82& 26.55& - &- &-\\
&RCAN$^{\dagger}$~\cite{zhang2018rcan} & 38.24 &	34.69 &	32.44
 & 34.02 &	30.59 &	28.81  &  32.35	 &29.29 &	27.73
 & 33.14 &	28.98 &	26.66
 & 39.37 &	34.44 &	31.26

\\
& NLSA${\dagger}$~\cite{NLSN_Mei_2021_CVPR}
&38.26 &	34.76 &	32.51
&34.11 &	30.68 &	28.89
& 32.35 &	29.30 & 	27.77 
&33.25 &	29.12 &	26.80
&39.43 &	34.55 &	31.42
\\
& SwinIR~\cite{liang2021swinir} &38.26 &34.77& 32.47& 34.14 &30.66& 28.85& 32.39 &29.31& 27.75& 33.29 &29.12& 26.76& 39.42&	34.58 &	31.34 
\\
& CAT-R-2$^{\dagger}$~\cite{cat_chen2022cross} & 38.30 &	34.74 &	32.40
&\textbf{34.21} &	30.68 &	28.83 
&32.40 & 29.29 & 27.72
&33.35 &	29.11 &	26.69
&39.49 &	34.52 &	31.17

\\
\cline{2-17}
&Baseline (Ours)
&38.29& 34.77 &	32.49 
&34.11 &30.68 &	28.84
& 32.40 &29.32 &27.74
&33.34 &29.12 &26.74
& 39.47 &34.53 & 31.28
\\
& TADT (Ours)
& \textbf{38.34} &\textbf{34.84}& \textbf{32.58} 
&34.13	&\textbf{30.75}	&\textbf{28.92} 
& \textbf{32.47} & \textbf{29.36} & \textbf{27.80}
&\textbf{33.50} &\textbf{29.32} & \textbf{26.96} 
& 
\textbf{39.57} &	\textbf{34.76} &\textbf{31.59}
\\
\hhline{-|-|---|---|---|---|---}
 \multirow{8}{*}{LIIF~\cite{2020liif}} & RDN~\cite{zhang2018rdn} & 38.17 & 34.68 & 32.50  &
33.97 & 30.53& 28.80 & 32.32& 29.26& 27.74& 32.87& 28.82& 26.68& 39.22 &	34.14 &	31.15
\\
&RCAN$^{\dagger}$~\cite{zhang2018rcan} & 38.21 &	34.74 &	32.59
& 34.02 &	30.61 &	28.89 
 & 32.36 &	29.29 &	27.77 
 &33.17 &29.03 &	26.86
& 39.37 &34.34 &31.31  

\\
&NLSA$^{\dagger}$~\cite{NLSN_Mei_2021_CVPR}& 38.30 &	34.86 & 32.73 
& 34.22 &	30.72 &	28.98
& 32.39 &	29.35 & 	27.83
&33.44 &	29.35 &	27.15
 & 39.58&34.67 &31.65

\\
& SwinIR~\cite{liang2021swinir} & 38.28 & 34.87 & 32.73  & 34.14 & 30.75& 28.98 &32.39 &29.34 &27.84 & 33.36& 29.33& 27.15& 39.53 & 34.65 &	31.67\\
& CAT-R-2$^{\dagger}$~\cite{cat_chen2022cross}&38.33 &34.91 &32.75 &34.27 &	30.79 &	29.02& 32.44 & 	29.38& 	27.86& 33.58 &29.44	&27.23& 39.53 &	34.66 &	31.69 

\\
\cline{2-17}
& Baseline (Ours)& 38.34 & 34.91 & 32.78 & 34.19 & 30.81 & 29.03 &32.44 &29.38 & 27.85 & 33.54 &29.49 &27.27  & 39.63	&34.74 &	31.77
\\
&TADT (Ours)& \textbf{38.38} & \textbf{34.96} & \textbf{32.83}
& \textbf{34.31}& \textbf{30.83} &	\textbf{29.07}
&\textbf{32.46} & 	\textbf{29.41} & 	\textbf{27.87}
&\textbf{33.65} &	\textbf{29.58}& \textbf{27.37}
 & \textbf{39.68} &\textbf{34.79} &	\textbf{31.83}
\\
 \hline
\multirow{8}{*}{LTE~\cite{lte-jaewon-lee}} &RDN~\cite{zhang2018rdn} &38.23 &34.72& 32.61 
&34.09& 30.58& 28.88 &32.36& 29.30& 27.77&33.04& 28.97& 26.81& 39.25 &34.28	&31.27\\
& RCAN$^{\dagger}$~\cite{zhang2018rcan}
  &38.24 &	34.77 &	32.60 & 34.04 &	30.64 &	28.87& 32.37 &	29.31 &	27.77 & 33.13 &29.04 &26.88 & 39.41 &34.39 &	31.30 \\
&NLSA$^{\dagger}$~\cite{NLSN_Mei_2021_CVPR}& 38.35 &	34.88 &	32.81
& 34.28	& 30.78 &	29.01
&32.43 & 29.39 &	27.86 
& 33.56 &29.43 &	27.25
& 39.64 &34.69 &	31.66

\\
& SwinIR~\cite{liang2021swinir} & 38.33 &34.89& 32.81 & 34.25& 30.80& 29.06 &32.44& 29.39& 27.86& 33.50& 29.41& 27.24& 
39.60 &	34.76 &	31.76 \\
& CAT-R-2$^{\dagger}$~\cite{cat_chen2022cross}& 38.36 &	34.91 &	32.80&
34.24 &	30.81 &	29.04 &32.47 &29.39	& 27.87 
& 33.60 &29.48 &27.27 
&39.61 &34.75 &	31.76
\\
\cline{2-17}
& Baseline (Ours)&38.39 &34.95&	\textbf{32.84} &34.25& 30.80& \textbf{29.06}&32.46 & 29.39 &27.86& 33.67&	29.51 &	27.33 & 39.66 &	34.77 &	31.77
\\
& TADT (Ours)& \textbf{38.42} & \textbf{34.99} &	32.83 & \textbf{34.37} &	\textbf{30.84} &	\textbf{29.06} & \textbf{32.47} &	\textbf{29.41} &	\textbf{27.88} & \textbf{33.70} & \textbf{29.57} & \textbf{27.36} & \textbf{39.72} &	\textbf{34.86} & \textbf{31.85} 
\\
\hline
\multicolumn{2}{r}{ArbSR~\cite{wang2021arbsr}} & 38.26 & 34.76  & 32.55 & 34.09 & 30.64 & 28.87 & 32.39 & 29.32 & 27.76 &33.14& 28.98 & 26.68 & 39.27 &34.55 & 31.36
 \\
 \multicolumn{2}{r}{LIRCAN~\cite{ijcai2023p63}} & 38.29 & 34.82 & 32.68& 34.33 & 30.77 & 28.97 &  32.42 & 29.36& 27.82& 33.13  & 29.11&  26.88 & 39.56 &34.77 & 31.71\\
 \multicolumn{2}{r}{EQSR~\cite{EQSR}} & 38.35 & 34.83 & 32.71 & 34.45 &30.82 & 29.12 & 32.46 &29.42 & 27.86 & 33.62 & 29.53 & 27.30 & 39.44 & 34.89 & 31.86
 \\
\hline
Upsampler & Feature Extractor  & $\times 6$ & $\times 8$ & $\times 12$ &  $\times 6$ & $\times 8$ & $\times 12$  & $\times 6$ & $\times 8$ & $\times 12$ &  $\times 6$ & $\times 8$ & $\times 12$ & $\times 6$ & $\times 8$ & $\times 12$ \\
 \hhline{-|-|---|---|---|---|---}
\multirow{8}{*}{MetaSR~\cite{hu2019meta}} & RDN~\cite{zhang2018rdn} & 29.04 &26.96& -& 26.51 &24.97 &- & 25.90 & 24.83 & -& 23.99 & 22.59&- &- & -  & -\\
 & RCAN$^{\dagger}$~\cite{zhang2018rcan} &
 29.02 &	26.97 &	24.63&
26.55 &	25.01 &	23.20 
& 25.91 &	24.83 &	23.47 &
24.06 &	22.65 &	21.05&
26.97 &	24.57 &	22.01
 \\
 
 &NLSA$^{\dagger}$~\cite{NLSN_Mei_2021_CVPR} 
 &29.07 &	27.00 &	24.72
 &26.56 &	25.07 &	23.25
&25.95 &	24.88 & 23.51 
& 24.20 &	22.78 &21.15
&27.11 &	24.71 &	22.13

\\
 &SwinIR~\cite{liang2021swinir}  & 29.09 & 27.02 &24.66  & 26.58& 25.09 & 23.23 &25.94 & 24.87 &23.52 & 24.16 &22.75 & 21.16 & 26.96 &	24.62 & 22.10
\\
 &CAT-R-2$^{\dagger}$~\cite{cat_chen2022cross} & 28.98 &	26.96 &	24.59 
 & 26.52 &	25.03 &	23.19 
 & 25.91 & 	24.85 &	23.48 
 & 24.11 &	22.73 &	21.12
& 26.86 &	24.54 &	22.06
\\
\cline{2-17}
& Baseline (Ours)& 29.08 &27.01 &	24.71
&26.56 & 25.07 &23.26
& 25.92 &24.85 & 23.50
& 24.14 &22.74 &21.12
& 26.88	&24.53 &22.00
\\
&TADT (Ours)& \textbf{29.16} &\textbf{27.11} &	\textbf{24.79} 
 & \textbf{26.65} &	\textbf{25.11} &	\textbf{23.23}
 & \textbf{25.97} &	\textbf{24.91} & \textbf{23.54}
 &\textbf{24.32} &	\textbf{22.91} & \textbf{21.24}
 &\textbf{27.20} &	\textbf{24.79} &	\textbf{22.20}
\\
 \hline
 \multirow{8}{*}{LIIF~\cite{2020liif}} & RDN~\cite{zhang2018rdn}  
 & 29.15 & 27.14 & 24.86 &26.64& 25.15& 23.24 &25.98& 
 24.91
 &
23.57
 & 24.20& 22.79 
 & 21.15 
 & 27.30 &	25.00 &	22.36
\\
&RCAN$^{\dagger}$~\cite{zhang2018rcan}
& 29.32 &	27.27 &	24.81
&26.69	&25.23 &	23.29
&26.01 &	24.95 & 	23.59 
& 24.35 &	22.92 &21.24 
&27.37 &	25.05 &	22.39

\\
&NLSA$^{\dagger}$~\cite{NLSN_Mei_2021_CVPR}&29.39 &27.24 &24.90 
&26.73 &	25.26 &23.34 
& 26.06 &	24.99 & 	23.62
&24.58 &	23.07 &	21.39
& 27.65 &	25.26 &22.53

\\
&SwinIR~\cite{liang2021swinir} & 29.46 & 27.36  &  24.98
&26.82& 25.34 & 23.37 
&26.07& 25.01&23.64
& 24.59& 23.14 & 21.43
& 27.66 &25.28 &22.57

 \\
&CAT-R-2$^{\dagger}$~\cite{cat_chen2022cross} & \textbf{29.53} &	\textbf{27.38}&	24.98
& 26.82 &	25.36 &	23.37
&26.09 &	25.02 &	23.62
&24.67 &	23.19 &	21.47
&27.72 &	25.31 &	22.58

\\
\cline{2-17}
&Baseline (Ours)& 29.45 &	27.34&	\textbf{25.03}
& 26.80 &	25.34 &	23.37
& 26.08 & 	25.03 &	23.64 
&24.68 &	23.22 &	21.51 
& 27.74 &	25.34 &	22.58
\\
&TADT (Ours)& 29.51 &	\textbf{27.38} & 25.01
&\textbf{26.84} &\textbf{25.34} &	\textbf{23.38}
&\textbf{26.10} &\textbf{25.05} &	\textbf{23.65}
&\textbf{24.75} &\textbf{23.27} & \textbf{21.53}
& \textbf{27.84}	&\textbf{25.39} &\textbf{22.59}
\\
\hline
 \multirow{8}{*}{LTE~\cite{lte-jaewon-lee}} & RDN~\cite{zhang2018rdn} & 29.32 & 27.26 &24.79 &  26.71 & 25.16 &23.31 & 26.01 &24.95&23.60 & 24.28&22.88	&21.22 
 & 27.46 &	25.09 &	22.43

\\
 & RCAN$^{\dagger}$~\cite{zhang2018rcan} &29.29 &	27.30 &	24.91 & 26.72 & 25.25 &	23.34 & 26.01 &	24.96 &	23.62 &
 24.33 &	22.92 &	21.29 & 27.44 &	25.09 &	22.43
\\
 &NLSA$^{\dagger}$~\cite{NLSN_Mei_2021_CVPR} & 29.43 &	27.33 &	25.02
&26.79 &25.32 &	22.36
 &26.08 &	25.02 &	23.65 
 &24.62 &	23.15 &	21.47
 &27.83 &	25.37 &	22.61
\\
 &SwinIR~\cite{liang2021swinir} & 29.50 & 27.35 &25.07 & 26.86 & 25.42 & \textbf{23.44} & 26.09 &25.03 &23.67 &24.62 &23.17&21.51 & 27.81 &25.39 &	22.65
\\
 &CAT-R-2$^{\dagger}$~\cite{cat_chen2022cross} & 29.41 &27.33 &24.96 & 26.85 &	25.34 &	23.40 & 26.09 & 	25.03 	& 23.65 & 24.68 & 23.21 & 21.50 & 27.84 &25.39 &22.66
\\
\cline{2-17}
 &Baseline (Ours)& 29.46 &27.39 &	25.04 & \textbf{26.86} &\textbf{25.36} &	23.41 & 26.09 & 	25.04 &	23.66
& 24.67 &	23.23 &	21.51 &
 27.85 &25.39 &	22.64
\\
&TADT (Ours)& \textbf{29.52} & \textbf{27.42} & \textbf{25.10} & 26.85 &25.35 &\textbf{23.44}& \textbf{26.11} &\textbf{25.05} &\textbf{23.67} & \textbf{24.72} &	\textbf{23.26} &	\textbf{21.54} & \textbf{27.93} &	\textbf{25.47}  &\textbf{22.70}\\
 \hline
  \multicolumn{2}{r}{ArbSR~\cite{wang2021arbsr}}& 28.45 &26.21&23.69 &26.22 & 24.55 &  22.55 &25.74 & 24.55 &23.07 & 23.70 & 22.13&20.40 & 26.18 & 23.58&  21.05\\
 \multicolumn{2}{r}{EQSR~\cite{EQSR}} & 29.41& - & - & 26.79 &- & - & 26.07 & - & - & 24.66 & - & - & 27.97& - & -
  \\
 \hline
\end{NiceTabular} 
}
\label{table:benchmark}
 \vspace{0mm}
\end{table*}
\clearpage

\begin{figure*}[!hp]
    \centering
    \vspace{-3mm}
\begin{overpic}
    [width=\linewidth]{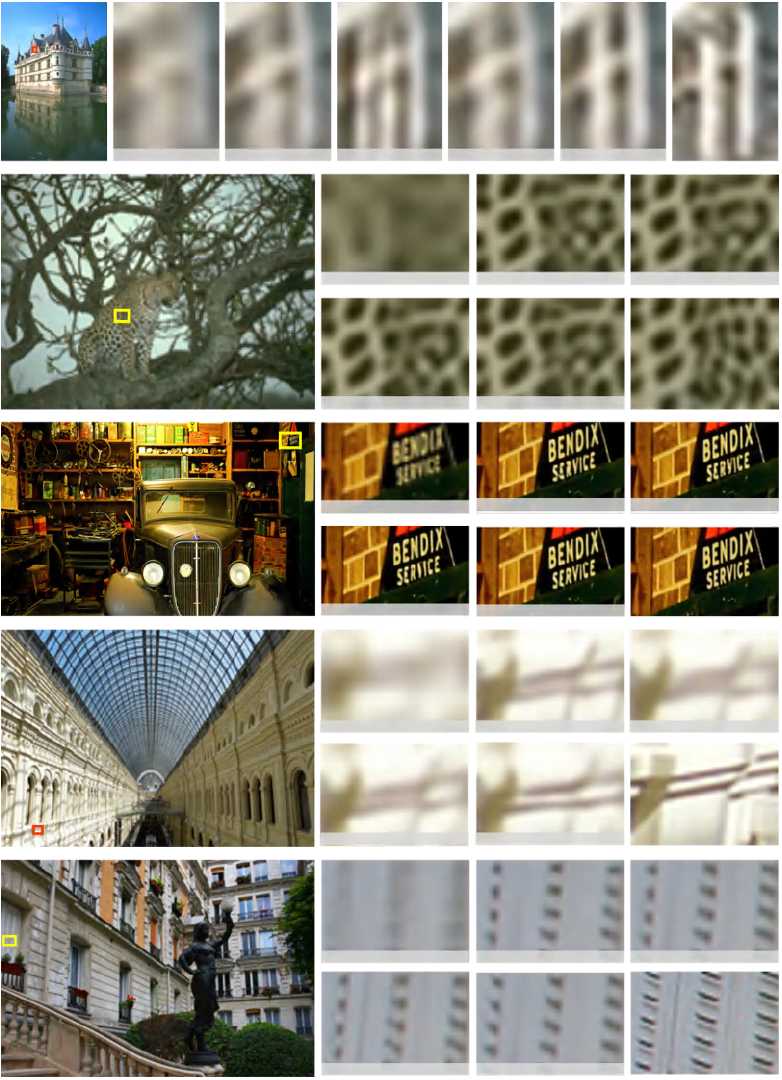}
 \put(2.5,84.3){\scriptsize B100, $s=2$}
 \put(14.7,84.3){\scriptsize LR}
     \put(12.1,85.4){\scriptsize PSNR / FLOPs}
 \put(21.1,85.4){\scriptsize 30.99 dB / 1628.48 G}
 \put(23.3,84.3){\scriptsize NLSA~\cite{NLSN_Mei_2021_CVPR}}

 \put(31.5,85.4){\scriptsize 30.83 dB / 475.28 G}
 \put(33.5,84.3){\scriptsize SwinIR~\cite{liang2021swinir}}

 \put(42,85.4){\scriptsize 31.03 dB / 477.52 G}
 \put(43.6,84.3){\scriptsize  CAT-R-2~\cite{cat_chen2022cross}}

 \put(52.7,85.4){\scriptsize 31.20 dB / 411.81 G}
 \put(54,84.3){\scriptsize TADT (Ours)}

 \put(66,84.3){\scriptsize GT}
%

\put(12, 61.2){\scriptsize B100, $s=2$}
\put(34.1,73.9){\scriptsize PSNR / FLOPs}
\put(36.2,72.8){\scriptsize LR}
 \put(46.6,73.9){\scriptsize 32.65 dB / 1628.48 G}
 \put(49,72.8){\scriptsize NLSA~\cite{NLSN_Mei_2021_CVPR}}

 \put(61,73.9){\scriptsize  32.78 dB / 475.28 G}
 \put(63,72.8){\scriptsize SwinIR~\cite{liang2021swinir}}

 \put(32.5,62.4){\scriptsize  32.79 dB / 477.52 G}
 \put(35,61.2){\scriptsize  CAT-R-2~\cite{cat_chen2022cross}}

 \put(46.6,62.4){\scriptsize  32.90 dB / 410.27 G}
 \put(49,61.2){\scriptsize  TADT (Ours)}

 \put(64,61.2){\scriptsize  GT}
%

\put(11.6, 42){\scriptsize DIV2K, $s=3$}
\put(34.1,52.88){\scriptsize  PSNR / FLOPs}
\put(36.2,51.6){\scriptsize LR}
\put(46.4,52.9){\scriptsize 31.40 dB / 11072.95 G}
 \put(49,51.6){\scriptsize NLSA~\cite{NLSN_Mei_2021_CVPR}}
\put(60.8,52.9){\scriptsize  31.44 dB / 3231.88 G}
 \put(63,51.6){\scriptsize SwinIR~\cite{liang2021swinir}}
\put(32.5,43.27){\scriptsize  31.44 dB / 3285.31 G}
 \put(35, 42.1){\scriptsize  CAT-R-2~\cite{cat_chen2022cross}}
 \put(46.6,43.26){\scriptsize  31.46 dB / 2498.31 G}
 \put(49,41.9){\scriptsize  TADT (Ours)}
 \put(64, 41.9){\scriptsize GT}

  \put(10.8, 20.7){\scriptsize Urban100, $s=4$}
  \put(34.1,32.55){\scriptsize PSNR / FLOPs}
\put(36.2,31.3){\scriptsize LR}
\put(46.4,32.55){\scriptsize 24.16 dB / 1845.75 G}
 \put(49,31.3){\scriptsize NLSA~\cite{NLSN_Mei_2021_CVPR}}
\put(60.8,32.55){\scriptsize  24.23 dB / 557.65 G}
 \put(63,31.3){\scriptsize SwinIR~\cite{liang2021swinir}}
\put(32.5,22){\scriptsize  24.18 dB / 560.29 G}
 \put(35, 20.7){\scriptsize  CAT-R-2~\cite{cat_chen2022cross}}
 \put(46.6,22){\scriptsize  24.27 dB / 505.00 G}
 \put(49,20.7){\scriptsize  TADT (Ours)}
 \put(64, 20.7){\scriptsize GT}

\put(10.8, -0.5){\scriptsize Urban100, $s=4$}
  \put(34.1,11.2){\scriptsize PSNR / FLOPs}
\put(36.2,10.2){\scriptsize LR}
\put(46.4,11.2){\scriptsize 22.06 dB / 1552.67 G}
 \put(49,10.2){\scriptsize NLSA~\cite{NLSN_Mei_2021_CVPR}}
\put(60.8,11.2){\scriptsize  22.18 dB / 456.26 G}
 \put(63,10.2){\scriptsize SwinIR~\cite{liang2021swinir}}
\put(32.5,0.7){\scriptsize  22.14 dB / 458.41 G}
 \put(35, -0.5){\scriptsize  CAT-R-2~\cite{cat_chen2022cross}}
 \put(46.6,0.7){\scriptsize  22.38 dB / 374.57 G}
 \put(49,-0.5){\scriptsize  TADT (Ours)}
 \put(64, -0.5){\scriptsize GT}
 
\end{overpic}
\vspace{0mm}
    \caption{ {
   \bfseries Visual comparison of feature extractors integrated with MetaSR~\cite{hu2019meta} on super-resolution natural images at scale 2, 3, 4.}  The highlighted regions are zoomed in for better view.
    } \label{fig:vis_comp1}
    \vspace{2mm}
\end{figure*}

\begin{figure*}[!h]
    \centering
    \vspace{-3mm}
\begin{overpic}
    [width=\linewidth]{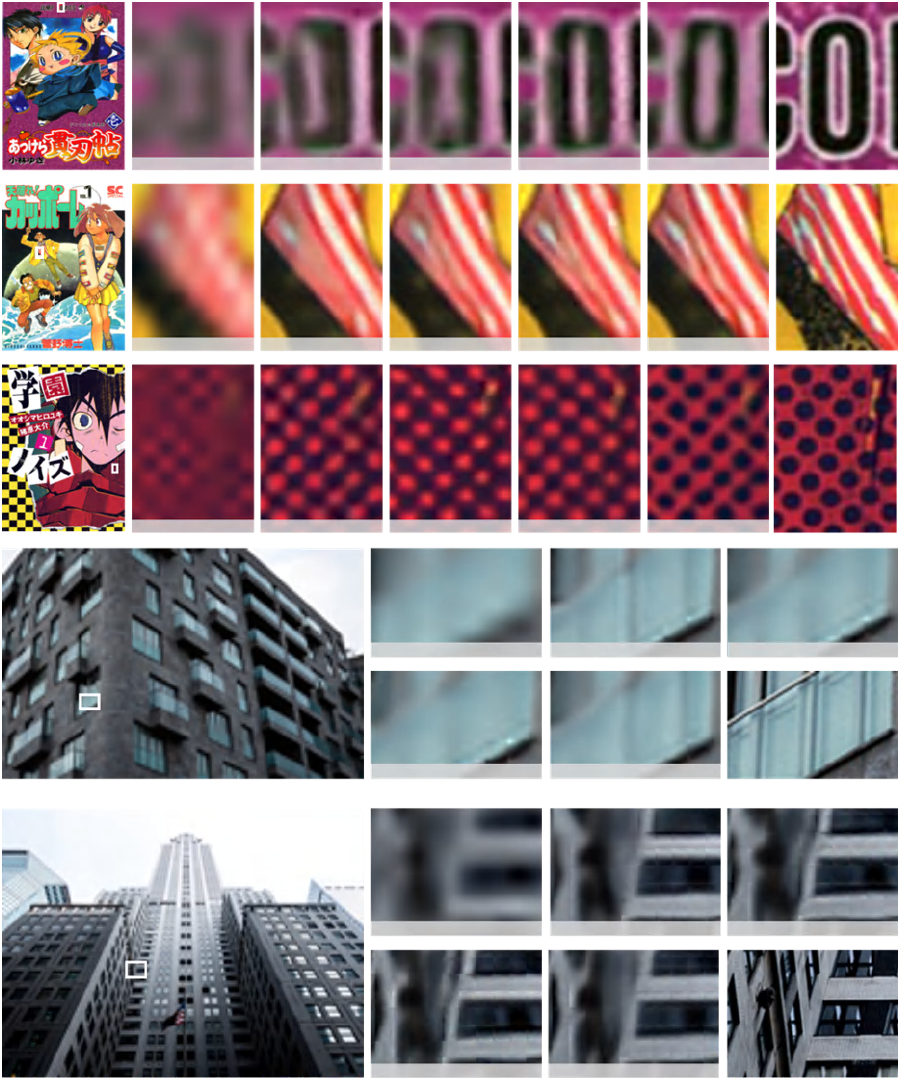}

 \put(2,83.3){\scriptsize  Manga109, $s=6$}
 \put(17.2,83.3){\scriptsize LR}
     \put(14.6,84.5){\scriptsize PSNR / FLOPs}

 \put(24.8,84.5){\scriptsize 25.23 dB / 1119.11 G}
 \put(28.3,83.3){\scriptsize NLSA~\cite{NLSN_Mei_2021_CVPR}}

 \put(37.1,84.5){\scriptsize 25.21 dB / 336.65 G}
 \put(39.5,83.3){\scriptsize SwinIR~\cite{liang2021swinir}}

 \put(48.8,84.5){\scriptsize 25.03 dB / 372.46 G}
 \put(51.5,83.3){\scriptsize  CAT-R-2~\cite{cat_chen2022cross}}

 \put(60.8,84.5){\scriptsize 25.31 dB / 312.38 G}
 \put(62.5,83.3){\scriptsize TADT (Ours)}

 \put(76,83.3){\scriptsize GT}
%
 \put(2,66.5){\scriptsize Manga109, $s=6$}
 \put(17.2,66.5){\scriptsize LR}
     \put(14.6,67.7){\scriptsize PSNR / FLOPs}
\put(24.8,67.7){\scriptsize 26.76 dB / 1119.11 G}
 \put(28.3,66.5){\scriptsize NLSA~\cite{NLSN_Mei_2021_CVPR}}
 \put(37.1,67.7){\scriptsize 26.75 dB / 336.65 G}
 \put(39.5,66.5){\scriptsize SwinIR~\cite{liang2021swinir}}
 \put(48.8,67.7){\scriptsize 26.61 dB / 372.46 G}
 \put(51.5,66.5){\scriptsize  CAT-R-2~\cite{cat_chen2022cross}}
\put(60.8,67.7){\scriptsize 26.90 dB / 315.67 G}
 \put(62.5,66.5){\scriptsize TADT (Ours)}
 \put(76,66.5){\scriptsize GT}
 
 \put(2,49.6){\scriptsize Manga109, $s=6$}
 \put(17.2,49.6){\scriptsize LR}
     \put(14.6,50.9){\scriptsize PSNR / FLOPs}
\put(24.8,50.9){\scriptsize 28.23 dB /  1119.11 G}
 \put(28.3,49.6){\scriptsize NLSA~\cite{NLSN_Mei_2021_CVPR}}
 \put(37.1,50.9){\scriptsize 28.17 dB / 336.65 G}
 \put(39.5,49.6){\scriptsize SwinIR~\cite{liang2021swinir}}
 \put(48.8,50.9){\scriptsize 28.10 dB /  372.46 G}
 \put(51.5,49.6){\scriptsize  CAT-R-2~\cite{cat_chen2022cross}}
\put(60.8,50.9){\scriptsize 28.79 dB / 315.67 G}
 \put(62.5,49.6){\scriptsize TADT (Ours)}
 \put(76,49.6){\scriptsize GT}

 \put(11,27){\scriptsize Urban100, $s=8$}
 \put(41,38.3){\scriptsize LR}
     \put(38.6,39.5){\scriptsize PSNR / FLOPs}
 \put(54.2,39.5){\scriptsize 24.83 dB /  434.21 G}
 \put(56.5,38.3){\scriptsize NLSA~\cite{NLSN_Mei_2021_CVPR}}
 \put(70,39.5){\scriptsize 24.71 dB / 126.74 G}
 \put(73,38.3){\scriptsize SwinIR~\cite{liang2021swinir}}
\put(38,28.2){\scriptsize 24.69 dB /  127.34 G}
 \put(40,27){\scriptsize  CAT-R-2~\cite{cat_chen2022cross}}
 \put(54.2,28.2){\scriptsize 24.87 dB / 125.67 G}
 \put(56,27){\scriptsize TADT (Ours)}
 \put(74.5,27){\scriptsize GT}
 
 \put(11.2,-1){\scriptsize DIV2K, $s=12$}
 \put(41,12.5){\scriptsize LR}
     \put(38.6,13.7){\scriptsize PSNR / FLOPs}
\put(54.2,13.7){\scriptsize 19.79 dB /  821.97 G}
 \put(56.5,12.5){\scriptsize NLSA~\cite{NLSN_Mei_2021_CVPR}}

 \put(70,13.7){\scriptsize 19.85 dB /  261.40 G}
 \put(73,12.5){\scriptsize SwinIR~\cite{liang2021swinir}}

\put(38,0.4){\scriptsize 19.78 dB / 280.14 G}
\put(40,-1){\scriptsize  CAT-R-2~\cite{cat_chen2022cross}}
 
 \put(54.2,0.4){\scriptsize 19.98 dB / 254.08 G}
 \put(56,-1){\scriptsize TADT (Ours)}
 \put(74.5,-1){\scriptsize GT}
\end{overpic}
\vspace{2mm}
    \caption{ {\bfseries Visual comparison of feature extractors integrated with MetaSR~\cite{hu2019meta} on super-resolution natural images at scale 6, 8, 12.}   The highlighted regions are zoomed in for better view.
    } 
    \label{fig:vis_comp2}
    \vspace{-4mm}
\end{figure*}

\begin{figure*}[!h]
    \centering
    \vspace{-3mm}
\begin{overpic}
    [width=\linewidth]{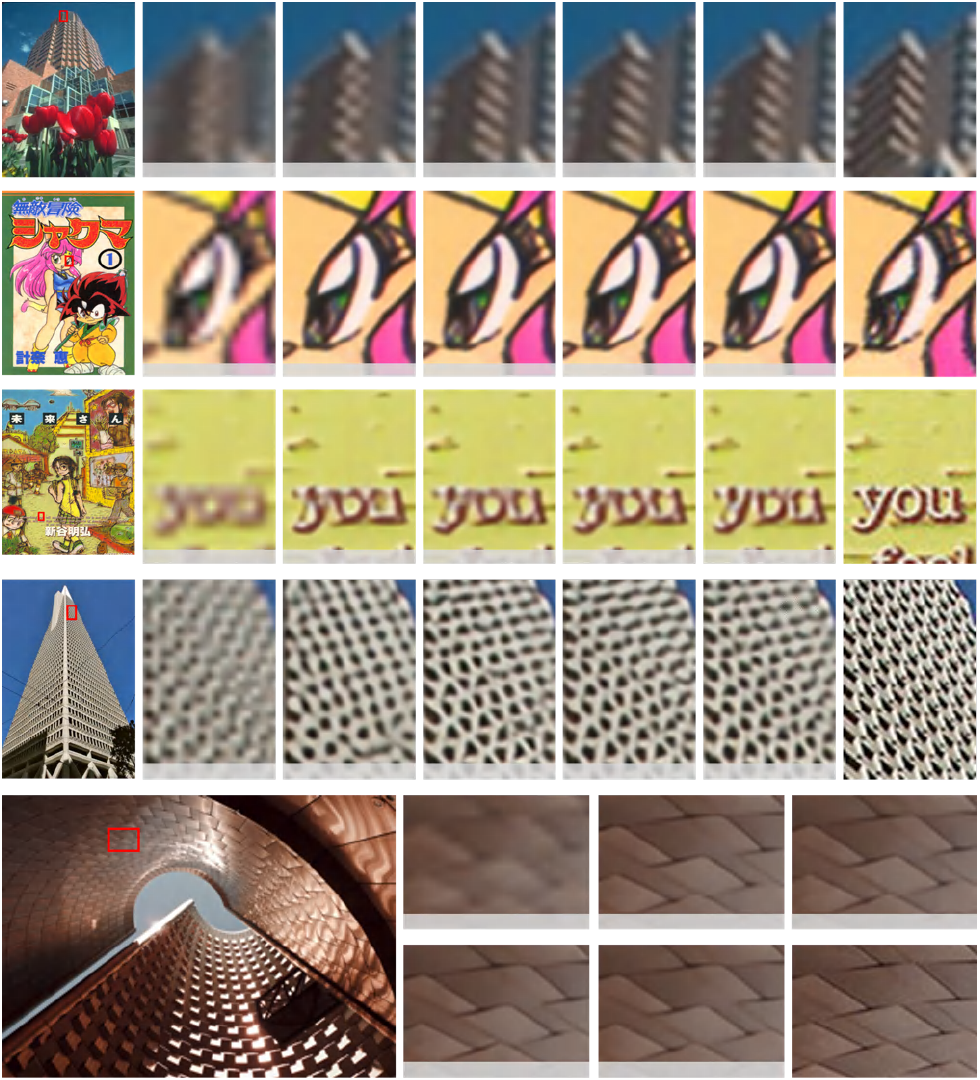}
    \put(0,84){\scriptsize }

 \put(3,82.6){\scriptsize B100, $s=2$}
 \put(18.5,82.6){\scriptsize LR}
     \put(15,83.85){\scriptsize PSNR / FLOPs}

 \put(26.8,83.85){\scriptsize 34.10 dB / 1628.48 G}
 \put(29.5,82.6){\scriptsize NLSA~\cite{NLSN_Mei_2021_CVPR}}

 \put(40.2,83.85){\scriptsize 34.11 dB / 475.28 G}
 \put(42.5,82.6){\scriptsize SwinIR~\cite{liang2021swinir}}

 \put(52.9,83.85){\scriptsize 34.39 dB / 477.52 G}
 \put(55, 82.6){\scriptsize  CAT-R-2~\cite{cat_chen2022cross}}

 \put(66,83.85){\scriptsize 34.70 dB / 418.00 G}
 \put(67.5,82.6){\scriptsize TADT (Ours)}
\put(83.5,82.6){\scriptsize GT}
%
\put(1.5,64.2){\scriptsize Manga109, $s=3$}
\put(18.5,64.2){\scriptsize LR}
     \put(15,65.4){\scriptsize PSNR / FLOPs}
\put(26.8,65.4){\scriptsize 37.23 dB /  4475.28 G}
 \put(29.5,64.2){\scriptsize NLSA~\cite{NLSN_Mei_2021_CVPR}}

 \put(39.8,65.4){\scriptsize 37.22 dB / 1319.68 G}
 \put(42.5,64.2){\scriptsize SwinIR~\cite{liang2021swinir}}

 \put(52.7,65.4){\scriptsize 37.34 dB / 1352.96 G}
 \put(55, 64.2){\scriptsize  CAT-R-2~\cite{cat_chen2022cross}}

 \put(65.5,65.4){\scriptsize 37.57 dB / 1141.51 G}
 \put(67.5,64.2){\scriptsize TADT (Ours)}
\put(83.5,64.2){\scriptsize GT}

 \put(1.8,26.8){\scriptsize Urban100, $s=4$}
\put(18.5,26.8){\scriptsize LR}
     \put(15,28.2){\scriptsize PSNR / FLOPs}
\put(26.8,28.2){\scriptsize 20.75 dB /  1856.30 G}
 \put(29.5,26.8){\scriptsize NLSA~\cite{NLSN_Mei_2021_CVPR}}

 \put(39.8,28.2){\scriptsize 21.22 dB / 557.66 G}
 \put(42.5,26.8){\scriptsize SwinIR~\cite{liang2021swinir}}

 \put(52.9,28.2){\scriptsize 21.47 dB / 560.29 G}
 \put(55, 26.8){\scriptsize  CAT-R-2~\cite{cat_chen2022cross}}

 \put(66,28.2){\scriptsize 22.06 dB / 508.01 G}
 \put(67.5,26.88){\scriptsize TADT (Ours)}
\put(83.5,26.8){\scriptsize GT}
\put(1.5,46.8){\scriptsize Manga109, $s=3$}
\put(18.5,46.8){\scriptsize LR}
     \put(15,48.1){\scriptsize PSNR / FLOPs}
\put(26.8,48.1){\scriptsize 25.31 dB /  4475.28 G}
 \put(29.5,46.8){\scriptsize NLSA~\cite{NLSN_Mei_2021_CVPR}}

 \put(39.8,48.1){\scriptsize 25.30 dB / 1319.68 G}
 \put(42.5,46.8){\scriptsize SwinIR~\cite{liang2021swinir}}

 \put(52.7,48.1){\scriptsize 25.29 dB / 1352.96 G}
 \put(55, 46.8){\scriptsize  CAT-R-2~\cite{cat_chen2022cross}}

 \put(65.5,48.1){\scriptsize 25.52 dB / 1148.75 G}
 \put(67.5,46.8){\scriptsize TADT (Ours)}
\put(83.5,46.8){\scriptsize GT}

\put(16.5,-1){\scriptsize Urban100, $s=4$}
\put(45,13){\scriptsize LR}
     \put(42.5,14.3){\scriptsize PSNR / FLOPs}
\put(58.5,14.3){\scriptsize 32.26 dB /  1964.95 G}
 \put(62,13){\scriptsize NLSA~\cite{NLSN_Mei_2021_CVPR}}

 \put(76.5,14.3){\scriptsize 32.43 dB /  583.01 G}
 \put(79,13){\scriptsize SwinIR~\cite{liang2021swinir}}

 \put(42,0.5){\scriptsize 32.56 dB / 611.22 G}
 \put(43.5, -1){\scriptsize  CAT-R-2~\cite{cat_chen2022cross}}

 \put(58.5,0.5){\scriptsize 32.99 dB / 507.42 G}
 \put(61,-1){\scriptsize TADT (Ours)}
\put(81,-1){\scriptsize GT}




\end{overpic}
\vspace{2mm}
    \caption{ {
   \bfseries Visual comparison of feature extractors integrated with LIIF~\cite{2020liif} on super-resolution for natural images at scale 2, 3, 4.}   The highlighted regions are zoomed in for better view.
    } \label{fig:vis_comp3}
    \vspace{-4mm}
\end{figure*}

\begin{figure*}[!h]
    \centering
    \vspace{-3mm}
\begin{overpic}
    [width=0.92\linewidth]{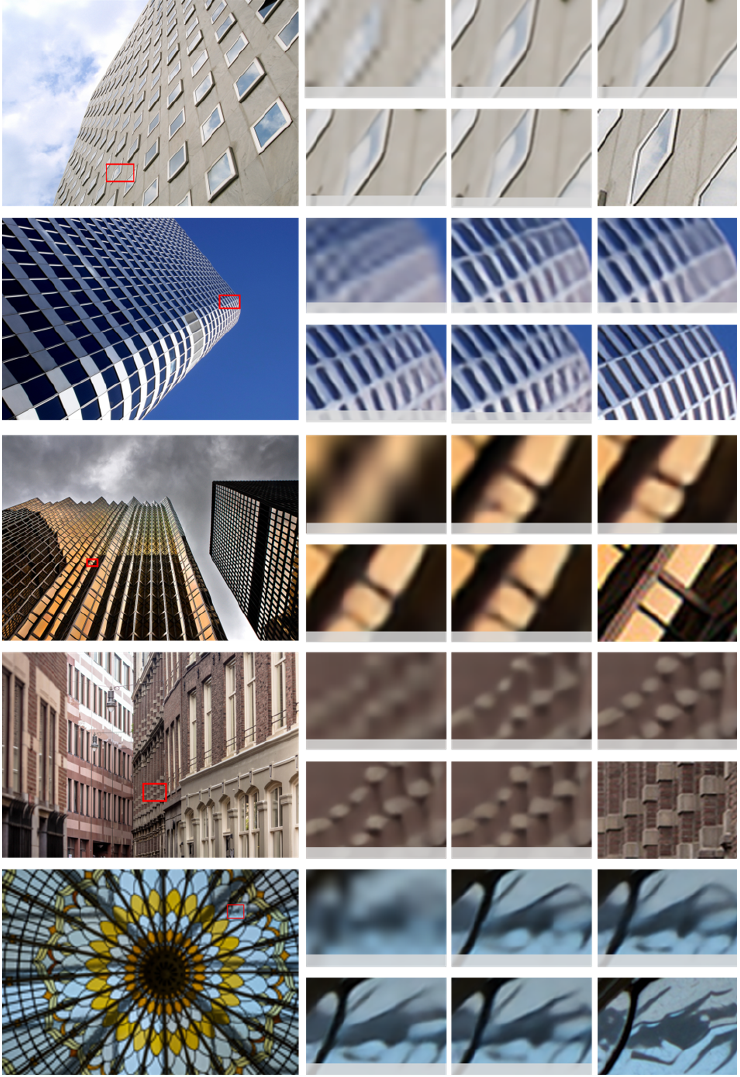}

\put(10.5,80.2){\scriptsize Urban100 $s=6$}
\put(33.5,90.2){\scriptsize LR}
     \put(31,91.1){\scriptsize PSNR / FLOPs}
\put(44,91.1){\scriptsize 25.65 dB /   980.57 G}
 \put(47,90.2){\scriptsize NLSA~\cite{NLSN_Mei_2021_CVPR}}

 \put(58,91.1){\scriptsize 25.55 dB /  296.26 G}
 \put(59.5,90.2){\scriptsize SwinIR~\cite{liang2021swinir}}

 \put(30.5,81){\scriptsize 25.87 dB / 315.16 G}
 \put(33, 80.1){\scriptsize  CAT-R-2~\cite{cat_chen2022cross}}

 \put(44,81){\scriptsize 26.19 dB / 254.08 G}
 \put(46,80.1){\scriptsize TADT (Ours)}
\put(61,80.1){\scriptsize GT}

\put(10.5,60.09){\scriptsize Urban100, $s=6$}
\put(33.5,70.1){\scriptsize LR}
     \put(31,71.05){\scriptsize PSNR / FLOPs}
\put(44,71.05){\scriptsize 26.41 dB /   923.09 G}
 \put(47,70.1){\scriptsize NLSA~\cite{NLSN_Mei_2021_CVPR}}

 \put(58,71.05){\scriptsize 26.90 dB /  278.83 G}
 \put(59.5,70.1){\scriptsize SwinIR~\cite{liang2021swinir}}

 \put(30.5,61.1){\scriptsize 26.89 dB / 280.14 G}
 \put(33, 60.09){\scriptsize  CAT-R-2~\cite{cat_chen2022cross}}

 \put(44,61.1){\scriptsize 27.19 dB / 254.08 G}
 \put(46,60.09){\scriptsize TADT (Ours)}
\put(61,60.09){\scriptsize GT}

\put(10.5,39.6){\scriptsize Urban100, $s=8$}
\put(33.5,49.55){\scriptsize LR}
     \put(31,50.55){\scriptsize PSNR / FLOPs}
\put(44,50.55){\scriptsize 18.61 dB /   461.18 G}
 \put(46,49.6){\scriptsize NLSA~\cite{NLSN_Mei_2021_CVPR}}

 \put(58,50.55){\scriptsize 19.02 dB /  139.41 G}
 \put(59.5,49.55){\scriptsize SwinIR~\cite{liang2021swinir}}

 \put(30.5,40.5){\scriptsize 19.11 dB / 152.81 G}
 \put(33, 39.585){\scriptsize  CAT-R-2~\cite{cat_chen2022cross}}

 \put(44,40.5){\scriptsize 19.34 dB / 127.11 G}
 \put(46,39.585){\scriptsize TADT (Ours)}
\put(61,39.585){\scriptsize GT}

\put(10.5,-0.6){\scriptsize DIV2K, $s=12$}
\put(33.5,9.5){\scriptsize LR}
     \put(31,10.5){\scriptsize PSNR / FLOPs}
    
\put(44,10.5){\scriptsize 27.36 dB /  814.83 G}
 \put(46,9.5){\scriptsize NLSA~\cite{NLSN_Mei_2021_CVPR}}

 \put(58,10.5){\scriptsize 27.44 dB /  261.40 G}
 \put(59.5,9.5){\scriptsize SwinIR~\cite{liang2021swinir}}

 \put(30.5,0.4){\scriptsize 27.57 dB / 280.14 G}
 \put(33,-0.6){\scriptsize  CAT-R-2~\cite{cat_chen2022cross}}

 \put(44,0.4){\scriptsize 27.60 dB / 254.08 G}
 \put(46,-0.6){\scriptsize TADT (Ours)}
\put(61,-0.6){\scriptsize GT}

\put(10.5,19.6){\scriptsize Urban100, $s=8$}
\put(33.5,29.45){\scriptsize LR}
     \put(31,30.35){\scriptsize PSNR / FLOPs}
    
\put(44,30.35){\scriptsize 23.50 dB /   461.18 G}
 \put(46,29.45){\scriptsize NLSA~\cite{NLSN_Mei_2021_CVPR}}

 \put(58,30.35){\scriptsize 23.64 dB /  139.41 G}
 \put(59.5,29.45){\scriptsize SwinIR~\cite{liang2021swinir}}

 \put(30.5,20.5){\scriptsize 23.72 dB / 152.81 G}
 \put(33, 19.6){\scriptsize  CAT-R-2~\cite{cat_chen2022cross}}

 \put(44,20.5){\scriptsize 23.90 dB / 127.11 G}
 \put(46,19.6){\scriptsize TADT (Ours)}
\put(61,19.6){\scriptsize GT}

\end{overpic}
\vspace{3mm}
    \caption{ {
   \bfseries Visual comparison of feature extractors integrated with LIIF~\cite{2020liif} on super-resolution for natural images at scale 6, 8, 12.} The highlighted regions are zoomed in for better view.
    } \label{fig:vis_comp4}
    \vspace{2mm}
\end{figure*}

\begin{figure*}[!h]
    \centering
    \vspace{-3mm}
\begin{overpic}
    [width=\linewidth]{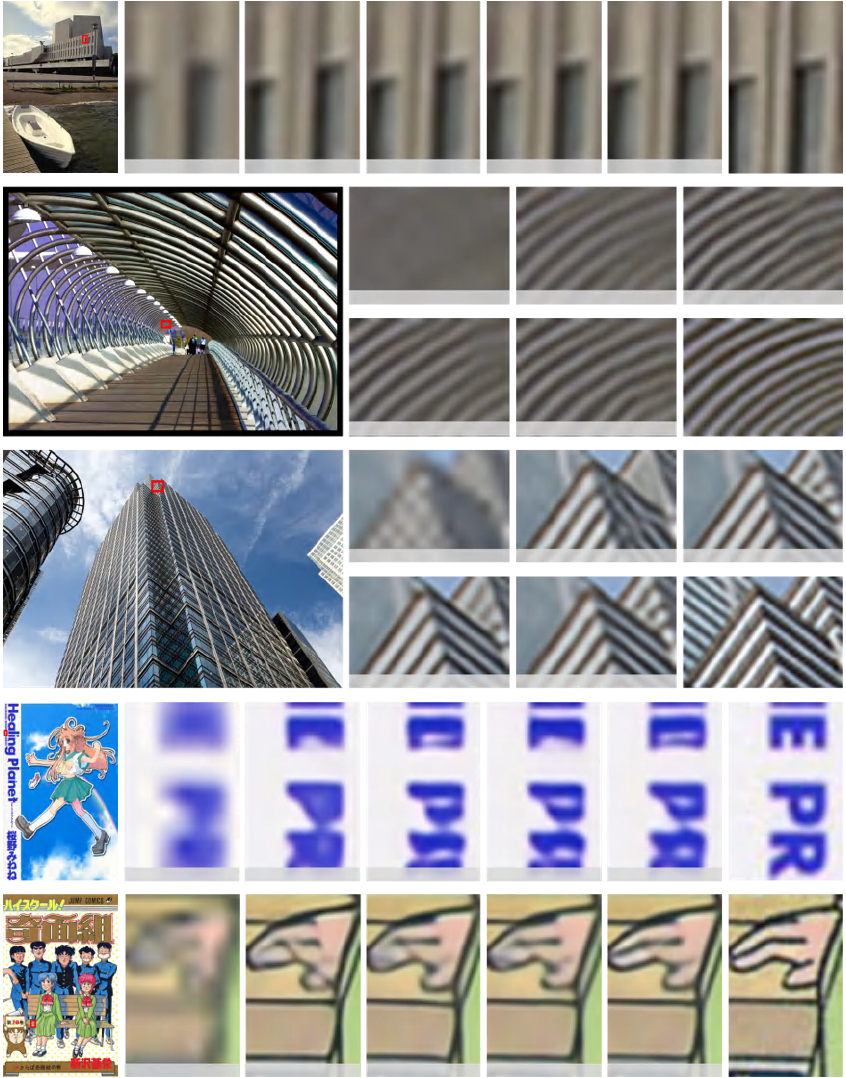}
    \put(2,83){\scriptsize B100, $s=2$}
 \put(15.8,83){\scriptsize LR}
     \put(13.5,84.25){\scriptsize PSNR / FLOPs}
    \put(23,84.25){\scriptsize 32.45 dB / 1628.48 G}
 \put(26,83){\scriptsize NLSA~\cite{NLSN_Mei_2021_CVPR}}

 \put(34.5,84.25){\scriptsize 32.87 dB / 475.28 G}
 \put(36.5,83){\scriptsize SwinIR~\cite{liang2021swinir}}

 \put(46,84.25){\scriptsize 32.84 dB / 477.52 G}
 \put(48, 83){\scriptsize  CAT-R-2~\cite{cat_chen2022cross}}

 \put(57,84.25){\scriptsize 32.99 dB / 416.66 G}
 \put(59,83){\scriptsize TADT (Ours)}
\put(72,83){\scriptsize GT}

   \put(11.5,58.75){\scriptsize Urban100, $s=3$}
 \put(39,70.8){\scriptsize LR}
     \put(36,72.1){\scriptsize PSNR / FLOPs}

   \put(51,72.1){\scriptsize 31.03 dB / 3268.28 G}
 \put(53,70.8){\scriptsize NLSA~\cite{NLSN_Mei_2021_CVPR}}

 \put(67,72.1){\scriptsize 30.59 dB / 987.78 G}
 \put(68.5,70.8){\scriptsize SwinIR~\cite{liang2021swinir}}

 \put(36,59.76){\scriptsize 30.96 dB / 1050.54 G}
 \put(38, 58.75){\scriptsize  CAT-R-2~\cite{cat_chen2022cross}}

 \put(51,59.76){\scriptsize 32.99 dB / 852.28 G}
 \put(52,58.75){\scriptsize TADT (Ours)}
\put(70,58.75){\scriptsize GT}

%
   \put(11.5,35.3){\scriptsize Urban100, $s=3$}
 \put(39,47){\scriptsize LR}
     \put(36,48){\scriptsize PSNR / FLOPs}

   \put(51,48){\scriptsize 24.90 dB /  2964.52 G}
 \put(53,47){\scriptsize NLSA~\cite{NLSN_Mei_2021_CVPR}}

 \put(67,48){\scriptsize 25.17 dB / 885.60 G}
 \put(68.5,47){\scriptsize SwinIR~\cite{liang2021swinir}}

 \put(36,36.4){\scriptsize 25.39 dB / 910.47 G}
 \put(38, 35.3){\scriptsize  CAT-R-2~\cite{cat_chen2022cross}}

 \put(52,36.4){\scriptsize 31.36 dB / 841.54 G}
 \put(54,35.3){\scriptsize TADT (Ours)}
\put(70, 35.3){\scriptsize GT}

\put(2.5,17.5){\scriptsize Manga109, $s=4$}

\put(16,17.3){\scriptsize LR}
     \put(13,18.7){\scriptsize PSNR / FLOPs}
   \put(23,18.7){\scriptsize 34.81 dB / 2517.60 G}
 \put(25,17.3){\scriptsize NLSA~\cite{NLSN_Mei_2021_CVPR}}

 \put(35,18.7){\scriptsize  34.77 dB / 762.03 G}
 \put(36.8,17.3){\scriptsize SwinIR~\cite{liang2021swinir}}

 \put(45.8,18.7){\scriptsize 32.84 dB / 786.31 G}
 \put(48, 17.3){\scriptsize  CAT-R-2~\cite{cat_chen2022cross}}

 \put(57,18.7){\scriptsize 35.22 dB / 737.23 G}
 \put(59,17.3){\scriptsize TADT (Ours)}
\put(72,17.3){\scriptsize GT}

\put(2.5,-1){\scriptsize Manga109, $s=4$}
\put(16,-1){\scriptsize LR}
     \put(13,0.45){\scriptsize PSNR / FLOPs}
   \put(23,0.45){\scriptsize 29.60 dB / 2517.60 G}
 \put(25,-1){\scriptsize NLSA~\cite{NLSN_Mei_2021_CVPR}}

 \put(35,0.45){\scriptsize  29.58 dB / 762.03 G}
 \put(36.8,-1){\scriptsize SwinIR~\cite{liang2021swinir}}

 \put(45.8,0.45){\scriptsize 29.65 dB / 786.31 G}
 \put(48, -1){\scriptsize  CAT-R-2~\cite{cat_chen2022cross}}

 \put(57,0.45){\scriptsize 29.92 dB / 745.76 G}
 \put(59,-1){\scriptsize TADT (Ours)}
\put(72,-1){\scriptsize GT}

\end{overpic}
\vspace{3mm}
    \caption{ {
   \bfseries Visual comparison of feature extractors integrated with LTE~\cite{lte-jaewon-lee} on super-resolution for natural images at scale 2, 3, 4.}  The highlighted regions are zoomed in for better view.
    } \label{fig:vis_comp5}
    \vspace{-4mm}
\end{figure*}
 
 \clearpage

\clearpage
\end{document}